\documentclass[11pt]{article}

\usepackage[final]{acl}

\usepackage{times}
\usepackage{latexsym}

\usepackage[T1]{fontenc}

\usepackage[utf8]{inputenc}

\usepackage{microtype}

\usepackage{inconsolata}

\usepackage{graphicx}
\usepackage{float}
\usepackage{enumitem}
\usepackage{xspace}
\usepackage{booktabs}
\usepackage{amsmath}
\usepackage{amssymb}
\usepackage{multirow}
\usepackage{xcolor}
\usepackage{array}
\usepackage{tcolorbox}
\usepackage{listings}
\tcbuselibrary{breakable}
\usepackage{caption}
\usepackage{stfloats}
\usepackage{pdflscape}

\tcbuselibrary{listings}

\definecolor{deltagreen}{RGB}{34,139,34}

\definecolor{deltared}{RGB}{180,0,0}

\DeclareRobustCommand{\ourmethod}{\textsc{ToolMaze}\xspace}

\title{When Tools Fail: Benchmarking Dynamic Replanning and Anomaly Recovery in LLM Agents}

\author{
Dongsheng Zhu$^{1}$\thanks{Equal contribution.}, \;
Xuchen Ma$^{2*}$, \;
Yucheng Shen$^{3}$, \;
Xiang Li$^{2}$, \;
Yukun Zhao$^{4}$, \\
\bfseries Shuaiqiang Wang$^{5}$, \;
Lingyong Yan$^{5}$\thanks{Corresponding author.}, \;
Dawei Yin$^{5}$ \\[4pt]
$^1$Shanghai AI Laboratory \quad
$^2$East China Normal University \\
$^3$Soochow University \quad
$^4$Shandong University \quad
$^5$Baidu Inc. \\[4pt]
\texttt{zhudongsheng@pjlab.org.cn} \quad
\texttt{xuchenma@stu.ecnu.edu.cn} \quad
\texttt{yanlingyong@baidu.com}
}

\begin{document}
\maketitle

\begin{abstract}
Existing benchmarks evaluate Tool-Integrated Reasoning (TIR) in LLMs on idealized ``happy paths'', largely overlooking real-world tool failures. 
We introduce \textsc{ToolMaze}, a benchmark for dynamic path discovery and error recovery in TIR agents. 
To separate systematic replanning from blind trial-and-error, \textsc{ToolMaze} adopts a two-dimensional design: DAG-based topological complexity and a $2 \times 2$ taxonomy of tool perturbations (explicit/implicit, transient/permanent). 
Evaluations show that perturbations degrade performance across nearly all models, with the sharpest drops under implicit semantic failures. Driven by systemic over-trust in corrupted outputs, Perturbation Recovery Rate (PRR) plummets by around 37\% in these scenarios, while complex topologies trap agents in futile trial-and-error loops. Crucially, agentic fault-tolerance improves with model scale $3.66\times$ slower than basic task execution, highlighting dynamic replanning as a distinct bottleneck unaddressed by model scaling or prompting. 
Data and code are available at \url{https://github.com/Zhudongsheng75/ToolMaze}.

\end{abstract}

\begin{figure}[!t]
  \centering
  \includegraphics[width=\linewidth]{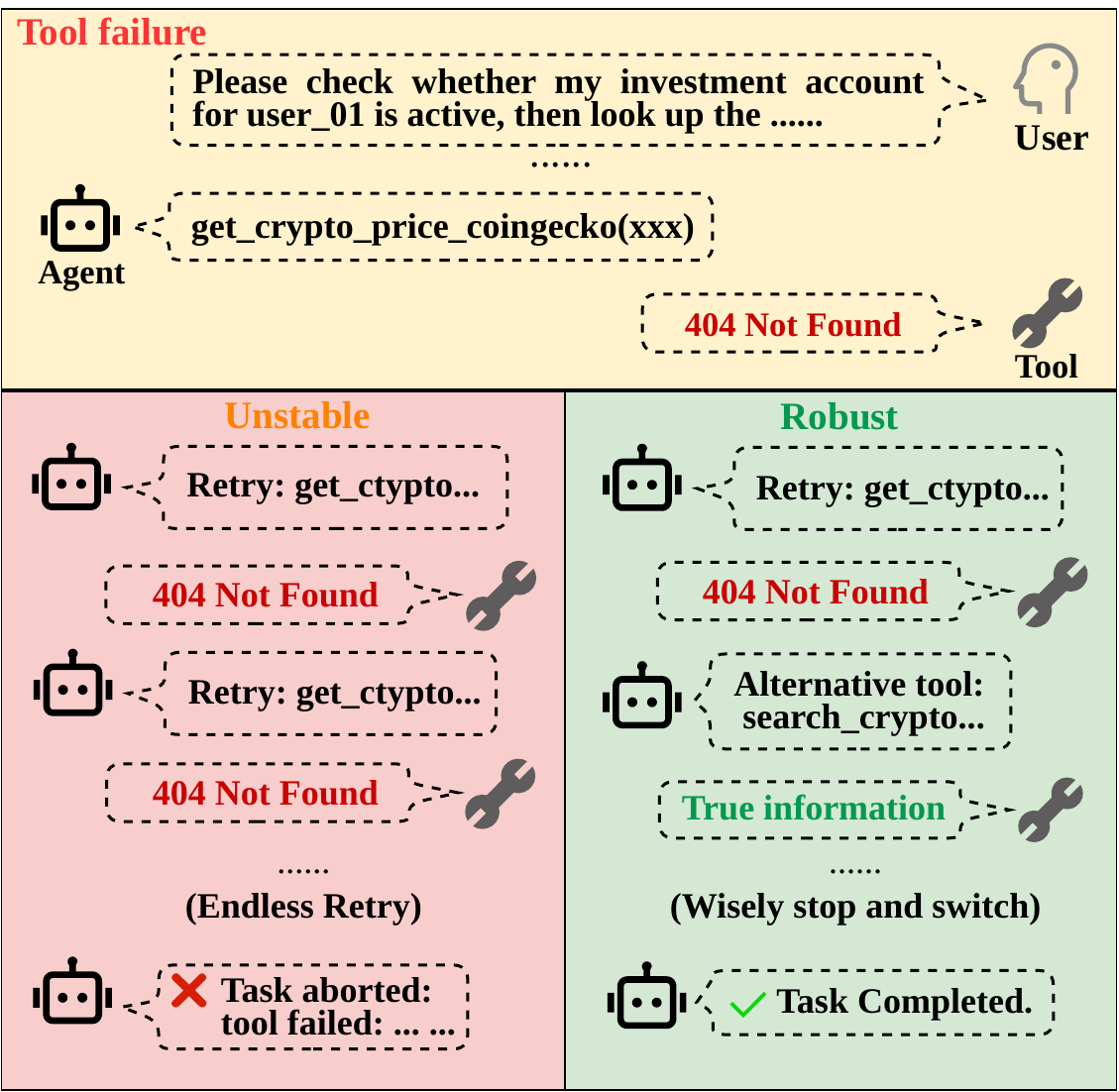}
  \caption{An illustrative example of agent behavior under tool failure. 
The \textcolor{orange}{\textbf{unstable}} agent aborts the task after an endless retry loop, whereas the \textcolor{green!60!black}{\textbf{robust}} agent wisely bypasses repeated failures by switching to an alternative tool.}
  \label{fig:example}
\end{figure}

\section{Introduction}

Integrating external tools has transformed LLMs from static knowledge repositories into Tool-Integrated Reasoning (TIR) agents~\cite{schick2023toolformer,qin2023toolllm}. 
However, prevailing benchmarks~\cite{li2023api,zhuang2023toolqa,guo2024stabletoolbench} evaluate these capabilities under a "happy path" fallacy, implicitly assuming perfectly stable and truthful environments.
Real-world tool execution, by contrast, is rarely a seamless linear pipeline. 
Instead, it forms a complex, failure-prone dependency graph~\cite{he2025sentinelagent}. Agents frequently encounter explicit failures such as network errors (e.g., 404, 429, timeout) that clearly block execution paths~\cite{zhang2026robust}. 
More insidiously, they face implicit failures~\cite{winston2025taxonomy, vuddanti2026paladin}—structurally valid but semantically corrupted responses, such as negative stock counts caused by delayed inventory updates. 
Without autonomous anomaly detection, agents blindly propagate these poisoned values, triggering cascading logic errors~\cite{romeo2026exploring, wingerter2025mitigating, alarcon2025explicating, kataria18intelligent}. 
This raises a crucial question: how resilient are LLM agents against such unpredictable instabilities and deceptive tool responses?

Ensuring system robustness necessitates a paradigm shift from linear execution to dynamic path discovery (Figure~\ref{fig:example}): when obstructed, an agent must smoothly transition from execution to exploration—detecting anomalies, backtracking, and systematically replanning—echoing the deliberate, slow-thinking style of System 2 reasoning~\cite{zhang2025system}. 
Current benchmarks~\cite{yao2024tau,backlund2025vending,yehudai2025survey} are structurally unequipped to measure these exploratory behaviors. 
Recent methods attempt to address the challenge of evaluating agent reliability under environmental variability by omitting failure injection~\cite{mohammadi2025evaluation}, artificially removing environmental variability~\cite{guo2024stabletoolbench}, or introducing fault/noise injection~\cite{gupta2026reliabilitybench,wang2026agentnoisebench,gurram2026evaluating}. However, they remain limited by two issues. First, they do not fully characterize the space of possible solutions, making it hard to tell whether an agent is systematically replanning or simply benefiting from tool substitutions~\cite{bean2026measuring}. Second, perturbations are introduced randomly rather than at pre-specified tool nodes. This makes it difficult to fairly measure agents' search efficiency or compare them reliably.

To systematically bridge these gaps, we introduce \textsc{\ourmethod}, a framework that shifts the evaluation of tool-using agents from static, single-trajectory execution to dynamic state-space exploration. 
The core idea of \textsc{\ourmethod} is to formulate robustness evaluation as a two-dimensional evaluation grid.
Every evaluation instance is situated at the intersection of two orthogonal axes: Topological Complexity ($\mathcal{C}$) and Perturbation Mode ($\mathcal{P}$). 
The $\mathcal{C}$-axis structures tasks on Directed Acyclic Graphs (DAGs) of increasing complexity ($\mathcal{C}1$--$\mathcal{C}4$) to precisely define available recovery paths. 
The $\mathcal{P}$-axis defines four perturbation modes by crossing two binary attributes: explicit versus implicit and transient versus permanent. Perturbations are injected at pre-specified tool nodes rather than randomly, enabling controlled evaluation of recovery behavior.
By generating instances conditioned on their $(\mathcal{C}, \mathcal{P})$ coordinates and exhaustively enumerating valid solutions, \textsc{\ourmethod} provides a complete ground truth set of recovery paths.
Furthermore, to rigorously quantify agent behavior, we move beyond binary Task Success Rate (TSR) by introducing Perturbation Recovery Rate (PRR) and Recovery Cost (RC). 
These metrics effectively isolate an agent's true replanning capability while strictly penalizing inefficient trial-and-error search.

Our three primary contributions are:
\begin{itemize}[leftmargin=1.2em, itemsep=2pt, topsep=2pt]
  \item \textbf{Novel Two-Dimensional Benchmark:}
        We introduce \textsc{\ourmethod}, the first benchmark to systematically evaluate dynamic path discovery and error recovery via orthogonal axes of DAG complexity and perturbation modes.

  \item \textbf{High-Quality, Scalable Synthesis Paradigm:}
        We design a scalable data synthesis paradigm that constructs DAG topologies prior to query naturalization, guaranteeing semantic coherence and enabling exhaustive solution enumeration.

  \item \textbf{Comprehensive Empirical Analysis:}
        Through experiments on state-of-the-art models, we show that agents often lack robust anomaly awareness, and that dynamic replanning captures a capability not reflected by general task success alone.
\end{itemize}

\section{Related Work}
\subsection{Complex Interactive Environments}
Early work establishes foundational tool-use abilities in LLMs \cite{guo2024stabletoolbench, li2023api, qin2023toolllm} while recent paradigms~\cite{lu2025toolsandbox, froger2025scaling, wang2025toolflow, wolflein2025llm, wijk2024re, cai2025llm} target stateful, open-ended environments.
Related settings also begin to stress robustness under evolving missions and disruptions: Multi-Mission Tool Bench\cite{yu2025multi} studies related and dynamic missions, while STT-Arena\cite{hui2026stt} evaluates replanning under spatio-temporal disruptions. 
Furthermore, planner-centric agents \cite{wei2026beyond} build global DAGs for multi-tool dependencies. Yet existing benchmarks rarely isolate robustness and recovery under noise, hallucinations, or execution failures.

\begin{figure*}[!t]
  \centering
  \makebox[\textwidth][c]{%
\includegraphics[width=1.02\textwidth]{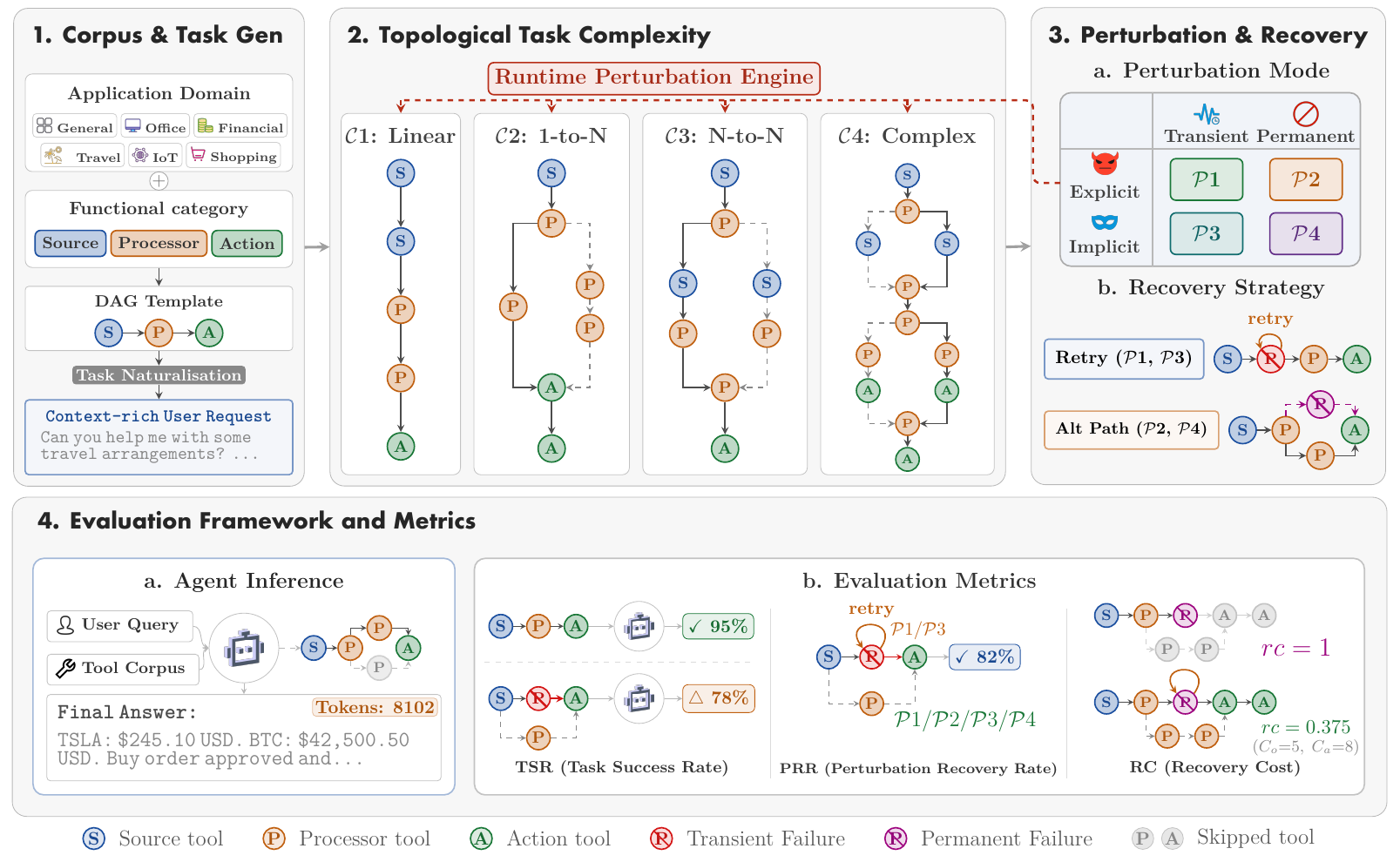}%
}
  \caption{Overview of the \textsc{ToolMaze} framework, illustrating its main components: (1) task generation from a curated tool corpus, (2) four levels of topological task complexity ($\mathcal{C}1$--$\mathcal{C}4$), (3) $2 \times 2$ taxonomy of perturbation modes ($\mathcal{P}1$--$\mathcal{P}4$), and (4) the evaluation framework with metrics including TSR, PRR, and RC.}
  \label{fig:overview}
\end{figure*}

\subsection{Robustness and Risk Evaluation}
The robustness of TIR agents has emerged as a critical research direction. Beyond adversarial injections, dynamic command generation \cite{zhang2025allies, jiang2025mimicking}, and manipulation of tool responses or selection mechanisms \cite{xiong2025more, sneh2025tooltweak}, recent benchmarks increasingly study reliability under realistic noise and failures. 
$\tau$-bench~\cite{yao2024tau} introduces pass$^k$ to distinguish consistent success from chance success. AgentNoiseBench~\cite{wang2026agentnoisebench} injects both user-noise and tool-noise, ReliabilityBench~\cite{gupta2026reliabilitybench} adopts chaos-engineering-style fault injection, and AgentProp-Bench~\cite{gurram2026evaluating} measures propagation cascades caused by parameter-level injections. ToolGym~\cite{xi2026toolgym} examines recovery from intermediate failures, highlighting how early errors can cascade through downstream tool interactions \cite{ruan2023identifying}.
Other methods improve resilience through architectural safeguards or recovery-oriented training \cite{xiang2025guardagent, xu2024reducing, zhang2026robust}. 
However, prior works still center on shallow chains, narrow attack surfaces, or unstructured failures, leaving implicit semantic failures and DAG-structured recovery underexplored.
\section{The \ourmethod Framework}
\label{sec:framework}

\ourmethod evaluates LLM agents along two orthogonal axes --- \emph{task complexity} ($\mathcal C$) and \emph{perturbation mode} ($\mathcal P$) --- whose cross product defines a structured evaluation space. As illustrated in Figure~\ref{fig:overview}, the framework is implemented as a pipeline consisting of task construction, perturbation engine, and evaluation.

\subsection{Tool Corpus}
\label{sec:corpus}

\ourmethod ships with a curated tool corpus of 270 tools, manually constructed based on real-world APIs and annotated with two complementary metadata attributes:

\noindent\textbf{Functional category.}
We distinguish three functional categories.
Source tools retrieve information from the simulated environment, for example \texttt{get\_weather} or \texttt{search\_news}.
Processor tools transform intermediate data with representative examples including \texttt{temperature\_converter} and \texttt{text\_translator}.
Action tools represent operations with external side effects and typically serve as terminal nodes, for example \texttt{send\_email} or \texttt{book\_flight}.
This three-way taxonomy imposes structural constraints during DAG construction. A well-formed task DAG must contain at least one Source node and one Action node, with Processor nodes and optional Action nodes bridging the two.

\noindent\textbf{Application domain.}
Tools are labeled with one of six domains: Financial, Travel, Office, Shopping, IoT, and General, covering a broad spectrum of realistic LLM tool-use scenarios from enterprise workflows to smart-home automation.
Domain labels constrain sampling during DAG construction to ensure semantic coherence. For example, a Travel-domain task will not chain a stock-price lookup with an email-sending action.
The framework further applies domain-balanced sampling so that generated tasks are distributed evenly across all six domains, preventing any single domain from dominating the benchmark.

\subsection{The \texorpdfstring{$\mathcal{C} \times \mathcal{P}$}{C x P} Evaluation Matrix}
\label{sec:cxp}

At the core of \ourmethod is a two-dimensional evaluation designed to systematically test an agent's fault tolerance and its ability to discover alternative valid tool-call paths after failures.
Rather than treating task generation and failure injection as isolated steps, every benchmark instance is fundamentally defined by a coordinate $(\mathcal C, \mathcal P)$ along two orthogonal axes: \textbf{Topological Task Complexity ($\mathcal C$)} and \textbf{Perturbation Mode ($\mathcal P$)}. 
Together, they dictate the available solution space (alternative tool-call paths and recovery strategies) and the nature of the encountered obstacle (error signal).

\paragraph{Axis 1: Topological Task Complexity ($\mathcal C$).} 
The $\mathcal C$-axis controls the DAG topology, determining how many alternative tool-call paths the agent may need to consider.
We define four progressive levels: $\mathcal C 1$ (linear) offers a single path with no alternatives, testing basic execution and graceful failure handling; $\mathcal C 2$ (1-to-N alternatives) introduces functionally equivalent substitutes, requiring direct single-step substitution; $\mathcal C 3$ (many-to-many multi-path) creates a combinatorial space of valid recovery paths across interacting sub-graphs, rigorously evaluating breadth-first planning; and $\mathcal C 4$ (integrated multi-branch) combines multiple $\mathcal C 2$ and $\mathcal C 3$ patterns within a single DAG, requiring the agent to reason over multiple branching nodes, each of which may contain 1-to-N or many-to-many recovery subgraphs.

\paragraph{Axis 2: Perturbation Mode ($\mathcal P$).} 

While the $\mathcal C$-axis defines \textit{where} an agent explores, the $\mathcal P$-axis dictates \textit{what} triggers this exploration. 
Beyond a Non-Perturbed (NP) baseline, tasks are evaluated under a $2 \times 2$ tool failure taxonomy governed by two orthogonal dimensions: error manifestation and temporal persistence. 
For anomaly detection, error manifestation fundamentally bifurcates into explicit and implicit failures. While explicit failures involve machine-readable exceptions (such as HTTP 404) that obstruct program execution, implicit failures generate structurally compliant but semantically flawed outputs, making autonomous verification indispensable.
For recovery strategies, temporal persistence distinguishes transient failures resolvable via simple retries from permanent ones that force dynamic rerouting or graceful termination. 
These dimensions yield four distinct modes ($\mathcal P 1$: Explicit-Transient, $\mathcal P 2$: Explicit-Permanent, $\mathcal P 3$: Implicit-Transient, $\mathcal P 4$: Implicit-Permanent) to comprehensively assess resilient replanning (details in Appendix \ref{sec:P1toP4}).

\subsection{DAG-Based Task Generation Pipeline}
\label{sec:task_pipeline}

To populate the $\mathcal C \times \mathcal P$ evaluation matrix defined in Section~\ref{sec:cxp}, our pipeline employs a tool-first paradigm, in which task graphs are constructed before natural-language queries, that guarantees both semantic coherence and the mathematical completeness of the ground-truth solution space.
An LLM-based \textit{DAG Architect} samples tools from the corpus (Section~\ref{sec:corpus}) to assemble a graph whose topology matches this target $\mathcal C$ level; solution-space enumeration then establishes a provably complete set of valid recovery paths before any natural-language query is generated.
The process consists of three sequential steps.

\paragraph{Step 1: DAG Assembly and Validation.}
For each target complexity level, the \textit{DAG Architect} assembles a DAG whose nodes are tool calls and whose edges encode data-flow dependencies.
Structural graph validation enforces acyclicity and confirms that the graph topology matches the intended $\mathcal C$ pattern.
Semantic validation further filters out incoherent data-flow compositions, such as cases in which a weather-object field is passed to a stock-price lookup tool.
To perform this check, a secondary LLM verifies data-flow coherence by checking whether each node's output fields can be meaningfully bound to its successors' input parameters.

\paragraph{Step 2: Solution-Space Enumeration.}

Rigorous fault-tolerance evaluation requires a mathematically complete space of valid recovery strategies. 
We establish \textit{substitutability relations} by clustering functionally equivalent tools or execution sub-graphs into schema-compatible alternative tool-call paths. 
These 1-to-N or many-to-many mappings instantiate the $\mathcal{C}2$ and $\mathcal{C}3$ complexity patterns. Traversing these alternatives enables the exhaustive enumeration of all valid topological DAG orderings. 
After pruning redundant cycles and suboptimal chains, we define the ground-truth solution space $\mathcal{S} = \{s_1, \dots, s_k\}$, designating the shortest sequence $s^* \in \mathcal{S}$ as the default path for the unperturbed baseline. 
Formalizing $\mathcal{S}$ allows \textsc{ToolMaze} to compare an agent's recovery trajectory against the optimal recovery path and quantify unnecessary tool calls.

\paragraph{Step 3: Task Naturalisation.}

To convert these DAGs into natural language task, a two-stage pipeline translates each DAG into a realistic user query. 
First, we distill the DAG into a task specification containing the core semantic arguments (e.g., \texttt{\{"goal": "stock price in EUR", "entity": "AAPL"\}}). 
Next, an LLM rewrites this skeleton into a context-rich request (Table~\ref{tab:prompt-task}), such as: ``\textit{I'm planning to invest in Apple---my account settles in euros. Can you check the current price for me?}''
To mitigate hallucinated content and omitted constraints during generation, we enforce a strict reverse validation step. 
An independent LLM reconstructs the tool dependencies solely from the generated query. The task is retained if and only if the reconstructed tool dependencies match the source DAG, minimizing semantic drift.

\subsection{Runtime Perturbation Engine}
\label{sec:perturbation_engine}

While Section \ref{sec:task_pipeline} generates the static task queries and ground-truth topological spaces, the Perturbation Engine is responsible for dynamically realizing the $\mathcal P$-axis during agent inference. 

\paragraph{Deterministic Perturbation Injection.} 
Each task carries a perturbation profile that specifies which tool on the preferred path $s^*$ should fail, and what synthetic response to return (e.g., an HTTP 404 error for $\mathcal P 1$/$\mathcal P 2$, or a structurally valid but semantically wrong output for $\mathcal P 3$/$\mathcal P 4$).
At runtime, when the agent calls a tool, the engine checks this profile: if the call matches a fault rule, the engine returns the pre-specified synthetic response; otherwise, it forwards the call to the standard tool simulator.
Because the injected responses are fixed per $(task, \mathcal P)$ pair, every model under evaluation receives identical injected fault responses, eliminating variance from the perturbation mechanism across runs.

\paragraph{Fault Activation Rules.}
In multi-path tasks, faults are associated with an alternative group instead of a fixed tool. Once the agent invokes any tool in the group, the engine assigns the fault to that chosen tool and disables further activations within the same group. This ensures that the perturbation is triggered regardless of which valid path the agent selects, while preventing multiple alternatives in the same group from being perturbed simultaneously. For $\mathcal C2$/$\mathcal C3$ tasks, this mechanism is applied globally to the single alternative group. For $\mathcal C4$ tasks, it is applied independently per parallel slot, so each branch can trigger at most one local perturbation without affecting the others. After activation, the $\mathcal P$-axis semantics apply uniformly: $\mathcal P1$/$\mathcal P3$ faults affect only the initial invocation, whereas $\mathcal P2$/$\mathcal P4$ faults make the targeted tool permanently unavailable or corrupted.

\subsection{Evaluation Framework and Metrics}
\label{sec:metrics}

To rigorously quantify agent robustness and dynamic path discovery, we evaluate execution trajectories over three complementary dimensions: overall completion (TSR), recovery capability (PRR), and replanning efficiency (RC).
Let $\mathcal{T}_m$ denote the set of evaluation trajectories under perturbation mode $m$. 
For each trajectory $\tau \in \mathcal{T}_m$, let $\mathbb{I}_{\text{succ}}(\tau) \in \{0, 1\}$ indicate successful task completion, and $\mathbb{I}_{\text{pert}}(\tau) \in \{0, 1\}$ indicate exposure to the injected perturbation.

\begin{table}[!t]
\centering
\small

\begin{tabular}{lcccc}
\toprule
\textbf{Metric} & \textbf{$\mathcal C 1$} & \textbf{$\mathcal C 2$} & \textbf{$\mathcal C 3$} & \textbf{$\mathcal C 4$} \\
\midrule
\multicolumn{5}{l}{\textit{Domain Involvement (out of 100 tasks per level)$^\dagger$}} \\
General   & 82 & 81 & 90 & 85 \\
Office    & 39 & 32 & 38 & 47 \\
Financial & 32 & 31 & 32 & 50 \\
Travel    & 33 & 30 & 32 & 47 \\
Shopping  & 31 & 29 & 31 & 49 \\
IoT       & 22 & 16 & 17 & 46 \\
\midrule
\multicolumn{5}{l}{\textit{Tool-Set Diversity}} \\
Mean Jaccard Sim. & 0.03 & 0.04 & 0.05 & 0.06 \\
\midrule
\multicolumn{5}{l}{\textit{Topological Statistics}} \\
Avg. Length & 5.30 & 5.75 & 6.66 & 11.00 \\
Length (Min/Max) & 5/7 & 5/9 & 6/10 & 8/15 \\
Valid Paths & 100 & 200 & 200 & 600 \\
\bottomrule
\multicolumn{5}{p{0.85\linewidth}}{\footnotesize $^\dagger$Column sums exceed 100 as cross-domain chaining is native to the benchmark.}
\end{tabular}
\caption{Domain distribution, tool-set similarity, and topological path statistics across $\mathcal C 1$-$\mathcal C 4$.}
\label{tab:task_stats}
\end{table}

\paragraph{Task Success Rate (TSR).} TSR captures the absolute task completion capability across the evaluation set:
\begin{equation}
\begin{split}
\text{TSR}_m &= \mathbb{E}_{\tau \sim \mathcal{T}_m}[\mathbb{I}_{\text{succ}}(\tau)] \\
&= \frac{1}{|\mathcal{T}_m|} \sum_{\tau \in \mathcal{T}_m} \mathbb{I}_{\text{succ}}(\tau)
\end{split}
\end{equation}
TSR reflects foundational tool-use proficiency (under non-perturbed) and overall resilience. Success under perturbation requires different strategies: retrying for transient faults ($\mathcal P 1$/$\mathcal P 3$), switching to alternative paths for permanent faults ($\mathcal C 2$--$\mathcal C 4$ under $\mathcal P 2$/$\mathcal P 4$), or executing a graceful termination when no alternatives exist ($\mathcal C 1$ under $\mathcal P 2$/$\mathcal P 4$).

\paragraph{Perturbation Recovery Rate (PRR).} To disentangle active recovery from passive avoidance, PRR evaluates the conditional probability of resolving an encountered perturbation:
\begin{equation}
\begin{split}
\text{PRR}_m &= \mathbb{P}(\text{Recovered} \mid \text{Perturbation}) \\
&= \frac{\sum_{\tau \in \mathcal{T}_m} \mathbb{I}_{\text{recov}}(\tau) \cdot \mathbb{I}_{\text{pert}}(\tau)}{\sum_{\tau \in \mathcal{T}_m} \mathbb{I}_{\text{pert}}(\tau)}
\end{split}
\end{equation}
where $\mathbb{I}_{\text{recov}}(\tau) = 1$ if the agent successfully executes a valid recovery strategy: (1) retrying for transient faults, (2) utilizing an alternative path, or (3) wisely aborting unsolvable tasks. PRR strictly evaluates error recovery independently of final task success.

\paragraph{Recovery Cost (RC).} RC quantifies replanning efficiency by penalizing unnecessary tool calls during recovery. 
For an exposed trajectory ($\mathbb{I}_{\text{pert}}(\tau) = 1$), let $c(\tau)$ and $c^*(\tau)$ be the empirical and theoretical minimum tool-call steps required from the first perturbed response to completion. 
The trajectory-level cost:

\begin{equation}
\mathcal{C}_{\text{rec}}(\tau) = 1 - \frac{c^*(\tau)}{\max\{c(\tau), c^*(\tau)\}}\cdot \mathbb{I}_{\text{succ}}(\tau) 
\end{equation}

\begin{table*}[!t]
\centering
\scriptsize
\setlength{\tabcolsep}{2.2pt}

\resizebox{\linewidth}{!}{%
\begin{tabular}{l|c|ccccc|ccccc|ccccc|c}
\toprule
\multicolumn{1}{c|}{\multirow{2}{*}{\textbf{Model}}}
& \multicolumn{1}{c|}{\textbf{TSR}\,(\%)\,$\uparrow$}
& \multicolumn{5}{c|}{\textbf{TSR}\,(\%)\,$\uparrow$}
& \multicolumn{5}{c|}{\textbf{PRR}\,(\%)\,$\uparrow$}
& \multicolumn{5}{c|}{\textbf{RC}\,(\%)\,$\downarrow$}
& \multicolumn{1}{c}{\multirow{2}{*}{\textbf{Avg.}\,$\uparrow$}} \\
\cmidrule(lr){2-2}\cmidrule(lr){3-7}\cmidrule(lr){8-12}\cmidrule(lr){13-17}
& NP
& $\mathcal P 1$ & $\mathcal P 2$ & $\mathcal P 3$ & $\mathcal P 4$ & Avg
& $\mathcal P 1$ & $\mathcal P 2$ & $\mathcal P 3$ & $\mathcal P 4$ & Avg
& $\mathcal P 1$ & $\mathcal P 2$ & $\mathcal P 3$ & $\mathcal P 4$ & Avg
& \\
\midrule
MiniMax-M2.7 (w/o hint) & 56.50 & 36.50 & 22.75 & 9.50 & 7.25 & 19.00 & 48.61 & 24.16 & 11.38 & 9.22 & 23.34 & 56.59 & 72.65 & 83.14 & 86.66 & 74.76 & 25.03 \\
MiniMax-M2.7 (w/ hint) & 55.50 & 44.25 & 25.25 & 9.25 & 9.00 & 21.94 & 57.66 & 27.75 & 9.86 & 12.05 & 26.83 & 50.40 & 70.71 & 85.43 & 85.34 & 72.97 & 27.50\rlap{$^{{\color{deltagreen}\scriptscriptstyle\uparrow}}$} \\
\midrule
Qwen3.5-35B-A3B (w/o hint) & 41.50 & 40.50 & 16.50 & 6.25 & 4.50 & 16.94 & 75.09 & 22.88 & 10.87 & 9.06 & 29.48 & 32.66 & 60.49 & 65.61 & 67.96 & 56.68 & 31.55 \\
Qwen3.5-35B-A3B (w/ hint) & 38.00 & 46.50 & 18.25 & 7.50 & 5.00 & 19.31 & 80.87 & 26.49 & 12.71 & 8.43 & 32.13 & 31.84 & 59.45 & 67.39 & 71.27 & 57.49 & 32.56\rlap{$^{{\color{deltagreen}\scriptscriptstyle\uparrow}}$} \\
\midrule
Qwen3.6-27B (w/o hint) & 42.25 & 42.50 & 15.50 & 4.75 & 4.75 & 16.88 & 77.78 & 21.38 & 7.85 & 7.97 & 28.75 & 36.97 & 66.74 & 70.43 & 72.84 & 61.74 & 29.65 \\
Qwen3.6-27B (w/ hint) & 44.50 & 48.50 & 27.75 & 11.25 & 8.25 & 23.94 & 80.15 & 33.06 & 16.55 & 11.52 & 35.32 & 32.98 & 63.69 & 72.40 & 74.83 & 60.98 & 34.13\rlap{$^{{\color{deltagreen}\scriptscriptstyle\uparrow}}$} \\
\midrule
Qwen3.5-397B-A17B (w/o hint) & 42.00 & 42.25 & 18.75 & 3.25 & 2.25 & 16.62 & 78.04 & 28.36 & 7.10 & 4.84 & 29.58 & 31.57 & 57.68 & 67.27 & 68.90 & 56.35 & 31.64 \\
Qwen3.5-397B-A17B (w/ hint) & 44.50 & 48.75 & 24.00 & 9.50 & 6.00 & 22.06 & 83.03 & 33.57 & 15.40 & 9.37 & 35.34 & 30.01 & 55.42 & 64.61 & 70.84 & 55.22 & 35.56\rlap{$^{{\color{deltagreen}\scriptscriptstyle\uparrow}}$} \\
\midrule
GLM-5.1 (w/o hint) & 75.75 & 60.00 & 32.50 & 8.00 & 8.50 & 27.25 & 67.19 & 35.33 & 9.04 & 10.15 & 30.43 & 37.86 & 66.47 & 88.65 & 89.11 & 70.52 & 32.28 \\
GLM-5.1 (w/ hint) & 74.75 & 66.25 & 37.00 & 14.75 & 17.00 & 33.75 & 75.22 & 38.72 & 15.65 & 18.35 & 36.98 & 32.32 & 62.48 & 81.98 & 81.24 & 64.50 & 38.14\rlap{$^{{\color{deltagreen}\scriptscriptstyle\uparrow}}$} \\
\midrule
Deepseek-V4-Pro (w/o hint) & 69.75 & 68.00 & 34.75 & 12.00 & 9.75 & 31.12 & 84.70 & 37.74 & 12.41 & 12.00 & 36.71 & 29.98 & 63.55 & 85.19 & 87.55 & 66.57 & 36.33 \\
Deepseek-V4-Pro (w/ hint) & 68.00 & 71.25 & \textbf{46.75} & 25.50 & 19.75 & 40.81 & 90.41 & 47.91 & 29.00 & 21.99 & 47.33 & 27.15 & 55.51 & 71.53 & 79.04 & 58.31 & 45.09\rlap{$^{{\color{deltagreen}\scriptscriptstyle\uparrow}}$} \\
\midrule
Claude-Sonnet-4-6 (w/o hint) & \textbf{77.00} & 49.00 & 27.00 & 5.75 & 6.00 & 21.94 & 59.68 & 27.68 & 5.08 & 7.08 & 24.88 & 48.95 & 71.56 & 90.88 & 91.75 & 75.79 & 27.35 \\
Claude-Sonnet-4-6 (w/ hint) & 76.25 & \textbf{75.50} & 39.50 & 27.75 & 25.25 & 42.00 & \textbf{90.61} & 38.94 & 29.99 & 26.63 & 46.54 & \textbf{23.22} & 61.85 & 69.58 & 73.29 & 56.99 & 46.14\rlap{$^{{\color{deltagreen}\scriptscriptstyle\uparrow}}$} \\
\midrule
GPT-5.5 (w/o hint) & 70.00 & 67.25 & 26.75 & 16.75 & 14.00 & 31.19 & 86.09 & 29.49 & 18.89 & 14.75 & 37.31 & 35.61 & 73.02 & 81.33 & 83.83 & 68.45 & 35.94 \\
GPT-5.5 (w/ hint) & 71.50 & 58.75 & 37.50 & \textbf{46.75} & 31.25 & 43.56 & 84.74 & 43.56 & \textbf{65.00} & 20.55 & 53.46 & 37.31 & 58.88 & \textbf{48.48} & 66.23 & 52.72 & 49.96\rlap{$^{{\color{deltagreen}\scriptscriptstyle\uparrow}}$} \\
\midrule
Gemini-3.1-Pro-Preview (w/o hint) & 71.25 & 59.75 & 30.00 & 6.75 & 9.00 & 26.38 & 73.49 & 33.52 & 8.09 & 10.84 & 31.48 & 37.53 & 67.64 & 88.64 & 87.24 & 70.26 & 32.19 \\
Gemini-3.1-Pro-Preview (w/ hint) & 68.75 & 65.25 & 42.50 & 43.25 & \textbf{34.00} & \textbf{46.25} & 90.21 & \textbf{53.09} & 54.98 & \textbf{29.36} & \textbf{56.91} & 29.73 & \textbf{52.59} & 49.38 & \textbf{63.53} & \textbf{48.81} & \textbf{52.95}\rlap{$^{{\color{deltagreen}\scriptscriptstyle\uparrow}}$} \\
\midrule
\bottomrule
\end{tabular}}
\caption{$\mathcal P$-based average results (\%) across perturbation modes. TSR (Non-Perturbed, NP) denotes the clean setting with no perturbation applied. Rows labelled \emph{(w/o hint)} use the standard prompt; \emph{(w/ hint)} rows use the failure-aware prompt. \textbf{Bold} marks the best value per column.}
\label{tab:main_results}
\end{table*}

The mode-level aggregated Recovery Cost averages the penalty across all runs:
\begin{equation}
\text{RC}_m = \frac{1}{|\mathcal{T}_m|} \sum_{\tau \in \mathcal{T}_m} \!\Big( \mathbb{I}_{\text{pert}}(\tau) \cdot \mathcal{C}_{\text{rec}}(\tau) \Big)
\end{equation}
Intuitively, $\text{RC} \to 0$ indicates optimal recovery ($c(\tau) = c^*(\tau)$), $\text{RC}$ increases toward 1 as the agent wastes steps on futile explorations, and $\text{RC} = 1$ constitutes an outright recovery failure.
\section{Experiments}
\label{sec:experiments}

\subsection{Experimental Setup}
\label{sec:setup}

\paragraph{Models.}
We evaluate representative open-weight and proprietary LLMs.
(1) Open-weight: GLM-5.1~\cite{glm5team2026glm5vibecodingagentic}, Deepseek-V4-Pro~\cite{deepseekai2026deepseekv4}, MiniMax-M2.7~\cite{minimaxm27}, Qwen3.5-35B-A3B~\cite{qwen3.5}, Qwen3.5-397B-A17B~\cite{qwen3.5}, Qwen3.6-27B~\cite{qwen3.6-27b}. (2) Proprietary: GPT-5.5~\cite{gpt55}, Gemini-3.1-Pro-Preview~\cite{gemini31}, Claude-Sonnet-4-6~\cite{claude}.

\begin{figure*}[!t]
    \centering
    \includegraphics[width=\linewidth]{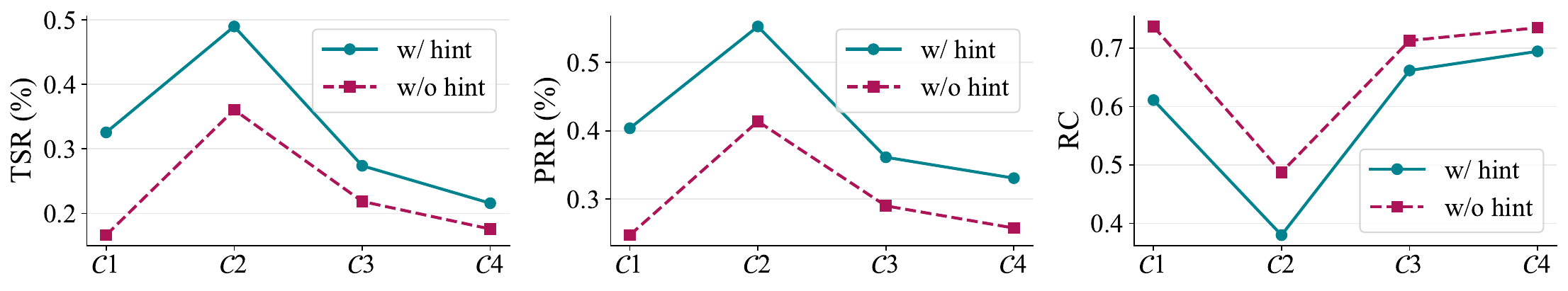}
    \caption{Mean metrics across complexity levels $\mathcal C 1$--$\mathcal C 4$, averaged over all evaluated models. Solid (dashed) lines correspond to the w/ (w/o) hint prompt.}
    \label{fig:plot_line_hint}
\end{figure*}

\paragraph{Dataset.}
\ourmethod builds upon a manually constructed corpus of 270 unique tools. 
Using this human-curated foundation, we leverage GPT-5.5 and Gemini-3.1-Pro-Preview to automatically synthesize 400 base tasks (100 each for $\mathcal C 1$–$\mathcal C 4$, prompts in Appendix~\ref{app:prompt}). 
Systematically applying Non-Perturbed baseline and our four perturbation modes expands this into 2,000 instances. Table~\ref{tab:task_stats} summarizes the task statistics and domain distribution. For more detailed dataset composition and diversity verification, please refer to Appendix~\ref{sec:appendix_dataset}.

\paragraph{Execution.}
Each task runs in a sandboxed environment with a maximum of 25 agent steps.
All LLM sampling uses temperature~1 and max tokens~16{,}000. 
To isolate the impact of prompting, we evaluate agents under two prompt configurations (Appendix~\ref{app:evaluation-prompt}). The first applies a standard tool-use prompt (w/o hint) across all modes (NP and $\mathcal P 1$--$\mathcal P 4$). The second substitutes a failure-aware prompt (w/ hint) specially for $\mathcal P 1$--$\mathcal P 4$, explicitly warning of potential tool failures and outlining recovery strategies.

\subsection{Main Results}
\label{sec:results}

Table \ref{tab:main_results} reports the primary evaluation results. 
To isolate the impact of different perturbation modes, the values presented for TSR, PRR, and RC are macro-averaged across all four topological complexity levels ($\mathcal C 1$--$\mathcal C 4$). 
Detailed breakdowns are provided in Appendix~\ref{sec:full_results}. 
To provide a holistic assessment, models are ranked by a composite score (\textbf{Avg.}), computed as the macro-average of each metric across all complexity-mode combinations: $(\text{Avg}(\text{TSR}) + \text{Avg}(\text{PRR}) + (1 - \text{Avg}(\text{RC}))) / 3$, using TSR for all settings and PRR/RC only for perturbation settings.
Overall, Gemini-3.1-Pro-Preview achieves the best comprehensive performance (\textbf{Avg.}), followed by GPT-5.5, Claude-Sonnet-4-6, and Deepseek-V4-Pro. 
Beyond leaderboard rankings, the results reveal two high-level findings:

\paragraph{Non-Perturbed vs.\ $\mathcal P 1$--$\mathcal P 4$.}
Almost all models exhibit substantial performance drops when transitioning from the unperturbed baseline (Non-Perturbed, NP) to perturbation modes ($\mathcal P 1$--$\mathcal P 4$). 
Even frontier models with strong NP performance (e.g., Claude-Sonnet-4-6 achieves 77.00\% TSR) exhibit substantial degradation under $\mathcal P1$--$\mathcal P4$, as reflected by lower TSR, PRR and higher RC.
This sharp contrast confirms that navigating linear ``happy paths'' and dynamically recovering from execution anomalies are decoupled capabilities.
The evaluation demonstrates that robustness does not naturally emerge from general instruction-following proficiency.

\paragraph{Standard vs.\ failure-aware prompt.}
Across all evaluated models, the \textit{w/ hint} configuration consistently outperforms its \textit{w/o hint} counterpart, yielding improvements ranging from +1.5\% to +20.8\%. 
This consistent performance gap highlights a systemic lack of intrinsic anomaly awareness in current agents. 
However, explicit prompting provides only partial mitigation.
Despite the perturbation hints provided in the failure-aware prompt, TSR under perturbation remains substantially lower than in the Non-Pert setting. 
Therefore, these observations indicate that current LLMs still lack robust dynamic replanning and anomaly recovery capabilities in tool-use and agentic tasks.

\begin{figure}[t]
    \centering
    \includegraphics[width=\linewidth]{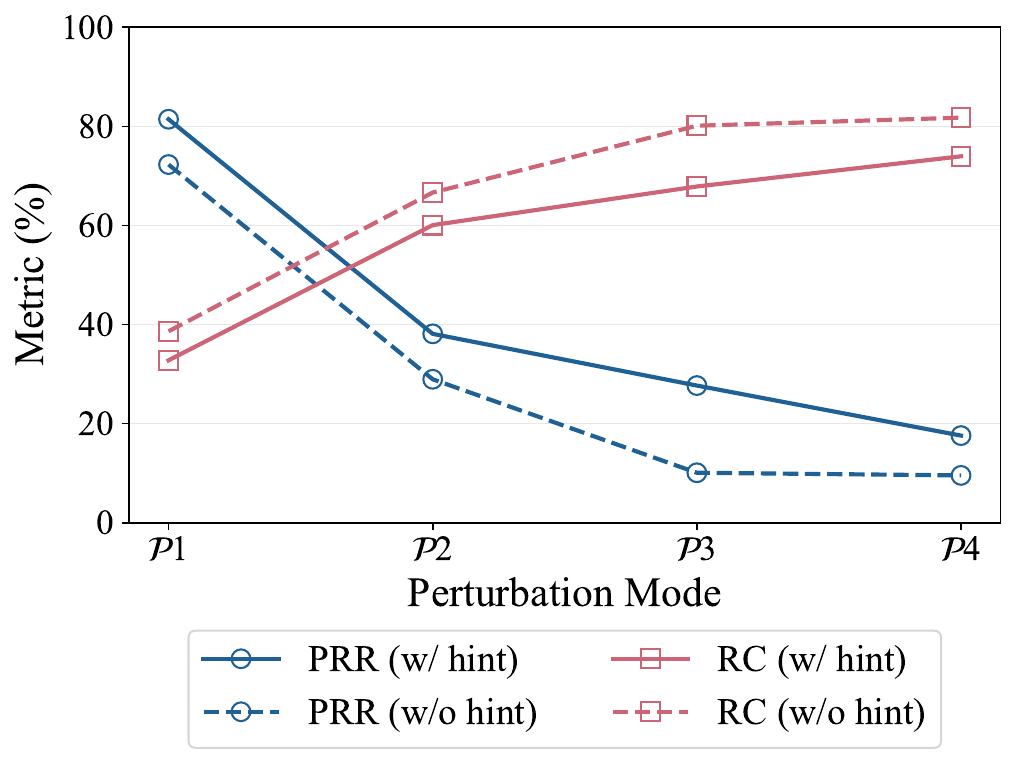}
    \caption{Mean PRR and RC across perturbation modes $\mathcal P 1$--$\mathcal P 4$, averaged over all evaluated models. Solid (dashed) lines correspond to the w/ (w/o) hint prompt.}
    \label{fig:PRR-RC}
\end{figure}

\subsection{Analysis}
\label{sec:analysis}

\paragraph{Effect of task complexity.}
As illustrated in Figure \ref{fig:plot_line_hint}, model resilience is strongest at $\mathcal C2$ rather than $\mathcal{C}$1: TSR and PRR reach their highest values, while RC reaches its lowest value.
$\mathcal{C}$1 tasks enforce a strict linear pipeline with zero structural redundancy. 
Consequently, any perturbation creates a single point of failure, requiring the agent to localize the anomaly despite having no alternative path.
In contrast, $\mathcal{C}$2 introduces alternative tool paths, enabling agents to bypass failed tools through rerouting.
However, as topological complexity scales to $\mathcal{C}$3 and $\mathcal{C}$4, performance deteriorates progressively. 
The challenge of navigating longer dependency chains and a combinatorially expanded search space quickly outstrips the benefits of path redundancy. 
As task complexity deepens, agents face greater difficulty, reflected in lower TSR and PRR, and make more unnecessary tool calls during recovery, reflected in higher RC.
Detailed results averaged across the perturbation dimension ($\mathcal{P}$) are provided in Appendix \ref{sec:full_results}.

\begin{figure}[t]
    \centering
    \includegraphics[width=\linewidth]{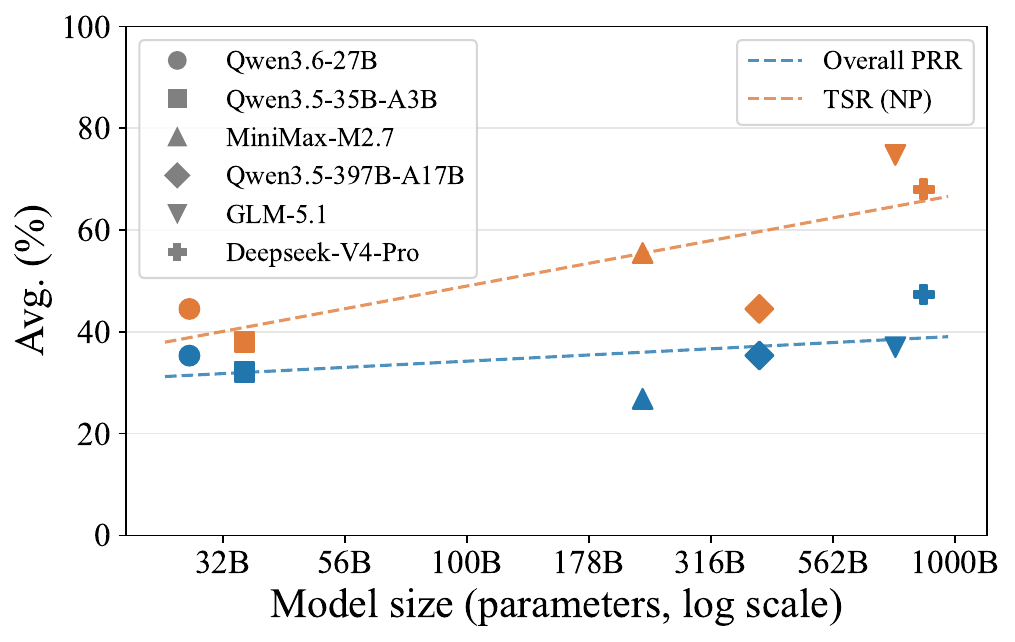}
    \caption{Overall PRR and TSR(NP) as a function of total model size for open-weight models (log scale).
             Dashed lines show linear fits in log-parameter space.}
    \label{fig:scaling}
\end{figure}

\paragraph{Opposing trends of PRR and RC.}

Figure~\ref{fig:PRR-RC} shows a clear inverse relationship between PRR and RC. As perturbation modes shift from easily detectable faults ($\mathcal P 1$) to deceptive, persistent errors ($\mathcal P 4$), models exhibit a monotonic decline in recovery success alongside a sharp increase in recovery cost. 
This indicates that current agents not only fail to resolve implicit semantic errors but also make many unnecessary tool calls during recovery.
The failure-aware prompt (\textit{w/ hint}) consistently outperforms the standard prompt (\textit{w/o hint}) by explicitly priming agents for potential anomalies, resulting in higher PRR and lower RC. This improvement, however, remains only a partial mitigation.
At $\mathcal P 1$, agents reliably detect and efficiently resolve explicit transient failures, achieving high PRR and low RC. 
Under implicit permanent failures ($\mathcal P4$), by contrast, PRR drops below 20\% while RC exceeds 70\%, suggesting that structurally valid but semantically incorrect outputs severely disrupt agent planning and lead to many unnecessary tool calls during recovery.

\paragraph{Fault-tolerance does not scale at the same rate as task completion.}
Figure~\ref{fig:scaling} plots overall PRR (averaged over $\mathcal C 1$--$\mathcal C 4$) and TSR(NP) against total model size (log scale) for the six open-weight models evaluated.
While both metrics improve with scale, their growth rates diverge significantly. 
A log-linear fit reveals that TSR(NP) grows by 17.85 percentage points (pp) per order of magnitude in parameter count, whereas PRR increases by merely 4.88 pp.
In other words, each order-of-magnitude increase in model size is associated with roughly $3.66\times$ more gain in baseline task completion than in fault-tolerance.
This stark discrepancy suggests that dynamic replanning and anomaly recovery do not emerge as natural by-products of general model scaling. 
Instead, they appear to represent a distinct capability that may require targeted training signals not yet systematically captured by current open-weight models.

\begin{figure}[t]
    \centering
    \includegraphics[width=\linewidth]{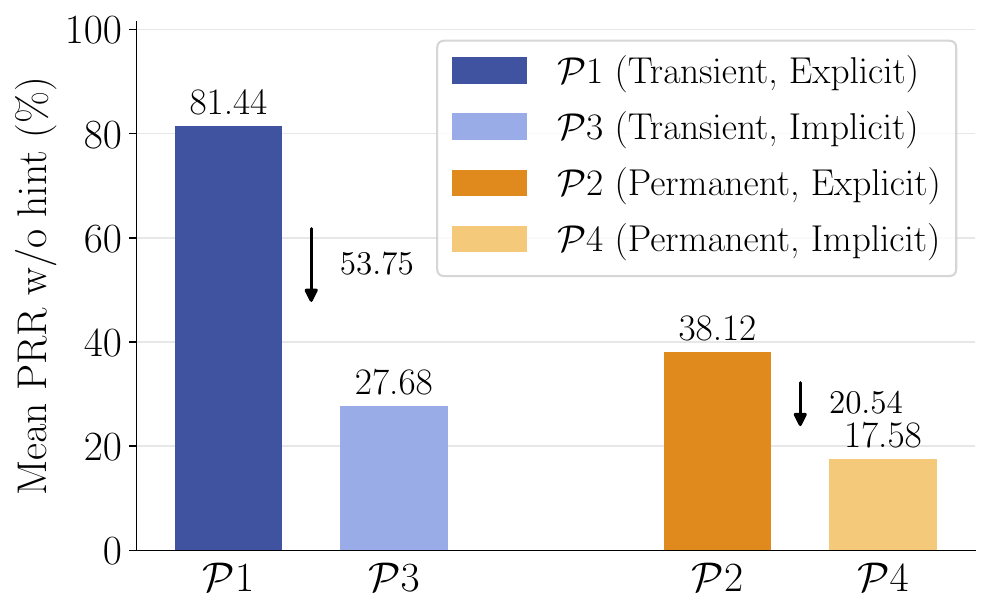}
    \caption{Average PRR performance (\textit{w/ hint}) between explicit perturbation and implicit perturbation.}
    \label{fig:prr_diff}
\end{figure}

\paragraph{The implicit–explicit trust gap.}
Current models exhibit over-trust in semantically perturbed tool outputs, exposing a profound vulnerability to implicit failures.
We measure this via the implicit--explicit PRR gap (i.e., $\text{PRR}(\mathcal P1) - \text{PRR}(\mathcal P3)$ for transient, and $\text{PRR}(\mathcal P2) - \text{PRR}(\mathcal P4)$ for permanent errors, both under the \textit{w/ hint} setting). 
As shown in Figure \ref{fig:prr_diff}, across all models, this gap remains strictly positive, averaging 37.15\% overall (53.75\% for transient and 20.54\% for permanent conditions).
This consistent drop indicates that detecting implicit anomalies remains a fundamental bottleneck for tool-use LLMs. 
Notably, the smaller gap in the permanent condition stems from a floor effect. 
Resolving persistent explicit errors ($\mathcal P 2$) is inherently difficult, demanding complex replanning ($\mathcal C 2$--$\mathcal C 4$) or graceful termination ($\mathcal C 1$). 
Because the PRR for $\mathcal P2$ is already low (38.12\% on average, compared with 81.44\% for $\mathcal P1$), there is less room for further degradation when failures become implicit under $\mathcal P4$.

\section{Conclusion}
\label{sec:conclusion}

We introduce \textsc{ToolMaze}, a DAG-based benchmark to systematically evaluate LLM agents' fault-tolerance and dynamic recovery, challenging the prevalent ``happy path'' evaluation fallacy. Our findings reveal a systemic lack of anomaly awareness in current models. Driven by over-trust in corrupted outputs, implicit semantic failures ($\mathcal{P}3/\mathcal{P}4$) cause recovery rates to plummet by 37.15\% compared to explicit errors, while deep topological complexity traps agents in futile trial-and-error loops. Crucially, agentic fault-tolerance improves $3.66\times$ more slowly with model scale than general task execution. This stark disparity proves that dynamic recovery is a distinct, foundational capability unaddressed by current scaling or superficial prompting strategies. Ultimately, \textsc{ToolMaze} underscores the urgent need to shift from linear execution toward deliberate, System 2-style autonomous anomaly detection and replanning for genuinely resilient agents.

\section*{Limitations}

While \textsc{\ourmethod} introduces a rigorous framework for evaluating dynamic replanning, it possesses certain limitations that provide avenues for future research: 

\paragraph{Evaluation in Controlled Topologies.} To provide a mathematically exact ground truth for recovery paths and rigorously penalize inefficient trial-and-error, \ourmethod prioritizes procedurally generated DAG topologies over completely open-ended web environments. While our reverse-validation pipeline ensures semantic realism within the benchmark, future work could extend our combinatorial task-generation paradigm to more unstructured, open-domain agentic workflows where success criteria are inherently ambiguous.

\paragraph{Extensibility of the Failure Taxonomy.} Our evaluation focuses on a foundational $2 \times 2$ taxonomy to systematically isolate the core dimensions of tool failures (explicit/implicit, transient/permanent). In real-world ecosystems, agents may also encounter compounded complexities, such as \textit{cascading failures} (where a silent error in one API triggers a chain reaction) or malicious adversarial injections. Building upon our foundational matrix to model these highly complex, multi-hop failure scenarios remains an exciting direction for advancing agentic security.



\bibliography{custom}

\appendix

\clearpage

\section{Perturbation Classes}
\label{sec:P1toP4}

We present defined perturbation types in Table \ref{tab:P1toP4}.
\begin{table*}[!t]
\centering

\begin{tcblisting}{
  listing only,
  colback=white!95!gray,
  colframe=gray!50!black,
  rounded corners,
  before skip=2pt,
  after skip=2pt,
  listing options={
    basicstyle=\ttfamily\fontsize{11pt}{10pt}\selectfont,
    breaklines=true,
    extendedchars=false,
    keepspaces=true,
    columns=fullflexible
  }
}
P1_ERROR_RESPONSES = [
  {"status": 503, "message": "Service Temporarily Unavailable", "error_type": "ServiceUnavailable"},
  {"status": 504, "message": "Gateway Timeout", "error_type": "Timeout"},
  {"status": 429, "message": "Too Many Requests", "error_type": "RateLimitExceeded"},
  {"status": 500, "message": "Internal Server Error", "error_type": "InternalServerError"},
  {"status": 502, "message": "Bad Gateway", "error_type": "BadGateway"},
  {"status": 409, "message": "Maximum Concurrent Connections Reached", "error_type": "ConcurrentLimit"},
  {"status": 409, "message": "Resource Temporarily Locked", "error_type": "ResourceLocked"},
  {"status": 503, "message": "Cache Unavailable", "error_type": "CacheUnavailable"},
  {"status": 502, "message": "Upstream Service Error", "error_type": "UpstreamError"},
]

P2_ERROR_RESPONSES = [
  {"status": 404, "message": "Tool Not Found", "error_type": "NotFound"},
  {"status": 410, "message": "Tool Permanently Removed", "error_type": "Gone"},
  {"status": 501, "message": "Not Implemented", "error_type": "NotImplemented"},
  {"status": 403, "message": "Access Forbidden", "error_type": "Forbidden"},
  {"status": 451, "message": "Unavailable For Legal Reasons", "error_type": "UnavailableForLegalReasons"},
  {"status": 402, "message": "Account Quota Exhausted", "error_type": "QuotaExhausted"},
  {"status": 410, "message": "API Version Deprecated", "error_type": "APIDeprecated"},
  {"status": 410, "message": "Feature Discontinued", "error_type": "FeatureDiscontinued"},
  {"status": 451, "message": "Service Unavailable In Your Region", "error_type": "GeoBlocked"},
  {"status": 403, "message": "Account Suspended", "error_type": "AccountSuspended"},
  {"status": 402, "message": "License Expired Or Invalid", "error_type": "LicenseInvalid"},
  {"status": 410, "message": "Data Source Retired", "error_type": "DataSourceRetired"},
]

P3_P4_CORRUPTION_TYPES = [
  "physics_violation",
  "tool_leakage",
  "irrelevant_response",
  "statistical_inconsistency",
  "internal_contradiction",
  "missing_required_fields",
  "logic_violation",
  "temporal_contradiction",
  "fact_contradiction",
]
\end{tcblisting}
\caption{Perturbation classes.}
\label{tab:P1toP4}
\end{table*}

\section{Detailed Dataset Statistics}
\label{sec:appendix_dataset}

This section provides a comprehensive statistical breakdown of the dataset corpus and the generated evaluation tasks. 

Table~\ref{tab:tool_stats} details the composition of the 270 simulated tools, categorized by their functional role and primary application domain. It also summarizes the distribution of the 126 alternative groups, which serve as the foundation for constructing multi-path topologies ($\mathcal C 2$--$\mathcal C 4$).

\begin{table}[H]
\centering
\small

\resizebox{\linewidth}{!}{%
\begin{tabular}{lr | lr | lr}
\toprule
\multicolumn{2}{c|}{\textbf{Functional Categories}} & \multicolumn{2}{c|}{\textbf{Domain Breakdown}} & \multicolumn{2}{c}{\textbf{Alternative Groups}} \\
\midrule
Action    & 123 & Office    & 61 & 1-to-1     & 44 \\
Source    & 78  & Financial & 58 & 1-to-$n$   & 26 \\
Processor & 69  & Travel    & 45 & $n$-to-$n$ & 56 \\
          &     & Shopping  & 36 &            &    \\
          &     & IoT       & 36 &            &    \\
          &     & General   & 34 &            &    \\
\midrule
\textbf{Total Tools} & \textbf{270} & \multicolumn{2}{c|}{} & \textbf{Total Groups} & \textbf{126} \\
\bottomrule
\end{tabular}
}

\caption{Overview of the tool corpus composition and alternative group structures.}
\label{tab:tool_stats}
\end{table}

Every tool in the corpus is a hand-crafted, deterministic simulation modelled after a real-world API.
Each implementation maps valid inputs to pre-defined outputs via static lookup tables, ensuring bit-exact reproducibility across evaluation runs while preserving realistic response schemas, parameter constraints, and error-handling conventions of the original services.
A subset of tools further maintains lightweight state through the shared execution context: for instance, \texttt{query\_availability} filters out time slots already consumed by earlier \texttt{schedule\_meeting} calls within the same trace, faithfully simulating the stateful behaviour of calendar APIs.
This simulation-based design eliminates external network dependencies, API-key management, and rate-limit variability, making benchmark results fully reproducible without sacrificing the structural fidelity needed to stress-test agent tool-use behaviour.

As introduced in Table~\ref{tab:task_stats}, domain footprints across the 400 base tasks are non-mutually exclusive. This occurs because complex tasks routinely chain APIs from different sectors (e.g., retrieving a flight schedule and subsequently recording the expense). To rigorously verify task diversity, we calculate the intra-category mean Jaccard similarity of constituent tool sets across all 4,950 possible task pairs within each 100-task subset. The consistently low similarity scores ($\le 0.06$) confirm high semantic variance and minimal template duplication.

\section{Prompt Templates}
\label{app:prompt}

\subsection{Evaluation Prompt}
\label{app:evaluation-prompt}
Table \ref{tab:prompt-evaluation-p0} and Table \ref{tab:prompt-evaluation} present the standard tool-use prompt and failure-aware prompt configurations, respectively. The standard tool-use prompt serves as a baseline, providing standard instructions for the agent to invoke tools in response to user queries. In contrast, the failure-aware prompt is a robustness-aware configuration that explicitly informs the agent of potential tool execution failures and outlines corresponding recovery strategies to mitigate these issues.

\begin{table}[H]
\centering
\footnotesize
\phantomsection

\begin{tcolorbox}[colback=white!95!gray,colframe=gray!50!black,rounded corners,label={scale-depression1}, title={Standard tool-use prompt.
}]

You are an autonomous AI assistant that executes tool-based workflows.

TASK COMPLETION

- The user query may describe multiple sub-tasks. You must complete ALL of them — do NOT skip any sub-tasks.

- Every fact, data retrieval, or state change must come from a tool call. Never substitute a tool result with internal knowledge, and never report a step as completed unless you actually called the corresponding tool and received its result in this conversation.

FINAL ANSWER FORMAT

- Be CONCISE and DIRECT. No emojis, markdown formatting (**, etc.), or bullet points.

- State only essential facts and requested results (e.g., confirmation codes). Omit internal technical details.
\end{tcolorbox}
\caption{Evaluation prompt for $\mathcal P 0$. It does not hint the possibility of tool failures or the corresponding recovery strategies.}
\label{tab:prompt-evaluation-p0}
\end{table}

\subsection{Task Description Generation Prompt}
\label{app:task-prompt}
Table \ref{tab:prompt-task} shows the prompt for task description generation from existing templates.

\begin{table}[!h]
\centering
\footnotesize
\phantomsection

\begin{tcolorbox}[colback=white!95!gray,colframe=gray!50!black,rounded corners,label={scale-depression}, title={Failure-aware prompt.
}]

You are an autonomous AI assistant that executes tool-based workflows. Tools may occasionally fail or return unexpected data.

TASK COMPLETION

- The user query may describe multiple sub-tasks. You must complete ALL of them — do NOT skip any sub-tasks.

- Every fact, data retrieval, or state change must come from a tool call. Never substitute a tool result with internal knowledge, and never report a step as completed unless you actually called the corresponding tool and received its result in this conversation.

- Sub-tasks may be accomplishable through different tool combinations.

FAULT HANDLING

- When evaluating tool outputs, consider whether the returned data is coherent and usable for the next step — fields should be present, values should be semantically reasonable, and types should match what the workflow requires.

- If you determine that a required step cannot be accomplished by any available tool or approach, stop the workflow and clearly report which step failed and why, rather than proceeding with incomplete results.

FINAL ANSWER FORMAT

- Be CONCISE and DIRECT. No emojis, markdown formatting (**, etc.), or bullet points.

- If SUCCESSFUL: State only essential facts and requested results (e.g., confirmation codes). Omit internal technical details.

- If HALTED: State clearly that the task was aborted and the reason (e.g., "Task aborted: the pricing tool returned corrupted data after retry and no other alternative approaches exist"). Do not apologize excessively.
\end{tcolorbox}
\caption{Evaluation prompt for $\mathcal P 1$-$\mathcal P 4$, which hints the possibility of tool failures or the corresponding recovery strategies.}
\label{tab:prompt-evaluation}
\end{table}
\begin{table}[!t]
\centering
\footnotesize
\phantomsection

\begin{tcolorbox}[colback=white!95!gray,colframe=gray!50!black,rounded corners,label={scale-depression2}, title={Task description generation.
}, fontupper=\fontsize{9pt}{8pt}\selectfont]

You are an expert data generation assistant.

You will be given a tool execution trace of a task in multiple messages (one step per message). Your job is to write a natural, colloquial user request (`task\_description`) that would reasonably lead to performing these steps.

**CRITICAL REQUIREMENTS - YOU MUST FOLLOW THESE EXACTLY**:

1. **MANDATORY COVERAGE**: The task description MUST require ALL steps in the execution trace, in order.

   - You will receive N steps. Your description MUST mention all N actions.
   
   - DO NOT skip any step, even if it seems redundant or obvious.
   
   - Each step must be explicitly or implicitly requested in the description.

2. **Step-by-Step Verification** (Internal checklist - verify before outputting):

   - Step 1: Does my description request this action? YES/NO
   
   - Step 2: Does my description request this action? YES/NO
   
   - Step 3: Does my description request this action? YES/NO
   
   - ... (for all N steps)
   
   - If ANY answer is NO, rewrite the description to include that step.

3. **FORBIDDEN - Past Tense / Completed Actions**:

    NEVER use: "I've already...", "I just...", "My ... is confirmed"
    
    NEVER imply actions are already done: "my booking", "the reservation I made"
    
    ALWAYS use future/imperative: "Please...", "I need to...", "Help me..."

4. **REQUIRED - Dependency Expression**:

   - If step B uses output from step A, use connecting words: "then", "after that", "and then"
   
   - Example: "Book a flight, then get the flight details, and send them to me" 
   
   - NOT: "Send me the flight details" (missing book step) 

5. **Natural Language**:

   - Do NOT mention tool names, schemas, or technical terms
   
   - Keep it realistic and concise (1-3 sentences)
   
   - Use the same language as the trace content
   
   - ONLY use facts from the trace blocks - do NOT invent details

6. **Verification Before Output**:

   - Count the steps in the trace: N steps
   
   - Count the actions in your description: M actions
   
   - If M < N, you MUST rewrite to include all steps
   
   - If M = N, verify each action maps to a step

domains: \{domains\}

language\_hint: \{language\_hint\}

You will receive the trace blocks next (one per message).

**REMEMBER**: Your description must request ALL \{num\_steps\} steps. Do not skip any step.

Output ONLY the user request text.

\end{tcolorbox}
\caption{Prompt for task description generation.}
\label{tab:prompt-task}
\end{table}

\begin{table*}[b]

\begin{tcblisting}{
    listing only,
    colback=white!95!gray,
    colframe=gray!50!black,
    rounded corners,
    listing options={
        basicstyle=\ttfamily\fontsize{10pt}{8pt}\selectfont,
        breaklines=true,
        extendedchars=false,
        keepspaces=true,
        columns=fullflexible
    }
}
### Role
You are a **Data Pipeline Architect** for C1 complexity level. Design a **5-7 step linear pipeline** using the provided tools.

### Pipeline Structure
Build a linear chain that starts from a Source-like step, contains meaningful Processor steps, and ends with an Action-like step. **No alternative nodes** — every step executes unconditionally.

### CRITICAL: Quality Requirements
**Every step must pass the Necessity Test**:
1. What NEW information does this step produce?
2. Which subsequent step(s) actually USE this information?
3. What would break if we removed this step?
**If you cannot answer all three questions clearly, DO NOT include that step.**

### Anti-Patterns to STRICTLY AVOID
1. **FORBIDDEN: Using tools on empty or meaningless data**
2. **FORBIDDEN: Type mismatches** — output type from step N must match input type of step N+1
3. **FORBIDDEN: Using `get_field_value` for simple data access** — only use it for complex nested structures
4. **FORBIDDEN: Fetching information that is never used by any subsequent step**

### Binding Reference Syntax
Every `arg_bindings` value must be exactly one of:
| Form | JSON | Meaning |
|------|------|---------|
| User input field | `{"source": "user_input", "field": "city"}` | A value the user provides |
| Field from prev step | `{"source": "prev", "field": "price"}` | One field extracted from the immediately previous step output |
| Whole prev output | `{"source": "prev"}` | Entire output dict of the previous step (use for object-typed params) |
| Field from earlier step | `{"source": "step", "step_number": 2, "field": "price"}` | One field from step 2's output |
| Whole earlier step | `{"source": "step", "step_number": 2}` | Entire output dict of step 2 |
| Constant | `{"source": "constant", "value": "USD"}` | Hard-coded literal |
**Do NOT** write raw strings or numbers directly as `arg_bindings` values.

### Type Compatibility Rules
Check the **Output Fields** and **Input Parameters** sections for each tool before writing bindings.

**Parameter expects `object`:**
- Use `{"source": "prev"}` or `{"source": "step", "step_number": N}` — no `field` key — to pass the entire previous output as the object.
- Only add `"field": "xxx"` if that specific field is itself an object (`dict`) in the upstream tool's Output Fields.
- **Never** point an object parameter at a scalar field (string, integer, boolean).

**Parameter expects `string` / `integer` / `number` / `boolean`:**
- Always include `"field": "xxx"` pointing to the exact scalar field name shown in the upstream tool's Output Fields.
- Use `user_input` if no upstream step produces the required scalar.

### Available Tools
{tool_specifications}

### Type Compatibility Checklist
Before setting each `arg_binding`, verify:
1. The previous step's output contains this field (check Output Fields)
2. The field's type matches the current parameter's expected type
3. The field's value is meaningful for the current operation
\end{tcblisting}
\caption{Prompt of $\mathcal C 1$ template construction (part 1).}
\label{tab:c1-template}
\end{table*}

\begin{table*}[b]

\begin{tcblisting}{
    listing only,
    colback=white!95!gray,
    colframe=gray!50!black,
    rounded corners,
    listing options={
        basicstyle=\ttfamily\fontsize{10pt}{8pt}\selectfont,
        breaklines=true,
        extendedchars=false,
        keepspaces=true,
        columns=fullflexible
    }
}
### Task
Select **5-7 tools** from the candidate list to create a coherent, useful task pipeline.
- Each step must produce NEW information or perform a meaningful transformation
- Each step's output must be USED by subsequent steps or the final result
- `user_input_schema` must cover every `user_input.xxx` field referenced in bindings

### Output Format (JSON only)
```json
{
  "success": true,
  "chain": [
    {
      "step": 1,
      "tool_name": "name_of_source_tool",
      "is_alternative": false,
      "reasoning": "Starts the chain by fetching X, which is needed by step 2 for Y",
      "arg_bindings": {
        "required_param": {"source": "user_input", "field": "required_param"}
      }
    },
    {
      "step": 2,
      "tool_name": "name_of_processor_tool",
      "is_alternative": false,
      "reasoning": "Processes data from step 1, transforming X into Y for step 3",
      "arg_bindings": {
        "data_param": {"source": "prev", "field": "output_field_from_step_1"},
        "config_param": {"source": "constant", "value": "some_literal"}
      }
    }
  ],
  "task_description": "One sentence describing the overall workflow",
  "user_input_schema": {
    "required_param": {"type": "string", "example": "example_value"}
  }
}
```

### Hard Rules
1. Chain must have exactly **5-7 steps**.
2. **No alternative nodes** — `is_alternative` must be `false` for every step.
3. **No duplicate tool names** in the chain.
4. Every required parameter must have a valid `arg_bindings` entry.
5. `user_input_schema` must cover every `user_input.xxx` field referenced in bindings.
6. If a valid chain cannot be constructed, return `{"success": false, "reason": "..."}`.

Respond ONLY with the JSON object, no additional text.
\end{tcblisting}
  \caption{Prompt of $\mathcal C 1$ template construction(part 2).}
  \label{tab:c1-template2}
\end{table*}

\begin{table*}[t]
  
\begin{tcblisting}{
    listing only,
    colback=white!95!gray,
    colframe=gray!50!black,
    rounded corners,
    listing options={
        basicstyle=\ttfamily\fontsize{10pt}{8pt}\selectfont,
        breaklines=true,
        extendedchars=false,
        keepspaces=true,
        columns=fullflexible
    }
}
### Role
You are a **Data Pipeline Architect** for C2 complexity level. Your goal is to design a data processing pipeline with **5-7 steps** that includes **at least one alternative node** using the provided tools and alternative groups.

### C2 Complexity Features
- **Chain Length**: 5-7 steps (flexible, not fixed like C1)
- **Alternative Nodes**: At least 1 step must be an alternative node (tools that can substitute each other)
- **Alternative Types**:
  - `one_to_one`: Two tools can substitute each other
  - `one_to_many`: One integrated tool OR multiple decomposed steps
- **Scope Restriction**: C2 only uses the two alternative types above. Other types such as `many_to_many` are reserved for C3/C4 and must not be used in C2.

### Data Flow Rules

1. **Flexible Data References**:
   - Step N can reference output from any previous step (step 1 to N-1)
   - Arguments can come from `user_input`, `constant`, `prev` (previous step), or `step` (any earlier step)
   - For `constant`, always use `{"source": "constant", "value": ...}` (NOT `field`)
   - For `step`, use `{"source": "step", "step_number": N, "field": "xxx"}` to reference step N's output

2. **Alternative Node Parameter Mapping**:
   - Use the parameter mapping defined in the alternative specification
   - Direct params: `{city: "city"}` - takes from input field named "city"
   - Constants: `{from_currency: "'USD'"}` - single quotes for string literals
   - References: `{a: "$stock_data.price"}` - $ prefix for previous tool outputs

3. **CRITICAL - Output Field Names**: When using `{"source": "prev", "field": "xxx"}`, use exact field names from Observed output keys.

4. **CRITICAL - Query vs Action Semantics**:
   - `get_*` / `query_*` / `check_*` tools return CURRENT state (for inspection)
   - `set_*` / `update_*` / `create_*` tools require TARGET state (desired outcome)
   - Do NOT use `prev` output from a query tool as input to its corresponding action tool
   - Example: `get_light_status` → `set_light_state` should use `user_input` for target state, NOT `prev` from query result
   - Dependencies between steps indicate execution order, NOT automatic parameter binding

5. **CRITICAL - Type Compatibility**:
   - **Object parameters**: If a parameter expects `type: object`, you MUST bind the entire object from `prev`, NOT individual fields
     - WRONG: `weather_data: {"source": "prev", "field": "temperature_celsius"}` (scalar field)
     - CORRECT: `weather_data: {"source": "prev", "field": "weather_result"}` (complete object)
   - **Scalar parameters**: If a parameter expects `type: string/number/integer`, bind a scalar field
   - Check the tool's parameter type specification before binding
   - For identifier parameters (`device_id`, `light_id`, `lock_id`, `media_id`, `target_device`), prefer `user_input` and avoid `prev` unless exact compatibility is guaranteed.
   - If previous step is an alternative node, `prev.field` must be valid for every alternative path (not just one path).
   - Use the "Alternative Output Field Hints" section: if an alternative group has `common keys across all paths: (none)`, do NOT use `source=prev` from that alternative step.
   - If previous step is an alternative node and you use `source=prev`, choose only from its listed `common keys across all paths`.
   - If unsure, keep the parameter from `user_input` instead of forcing a `prev` link.
\end{tcblisting}
\caption{Prompt of $\mathcal C 2$ template construction (part 1).}
  \label{tab:c2-template}
\end{table*}

\begin{table*}[t]
  
\begin{tcblisting}{
    listing only,
    colback=white!95!gray,
    colframe=gray!50!black,
    rounded corners,
    listing options={
        basicstyle=\ttfamily\fontsize{10pt}{8pt}\selectfont,
        breaklines=true,
        extendedchars=false,
        keepspaces=true,
        columns=fullflexible
    }
}
### Available Tools

{tool_specifications}

{alternative_specifications}

### Task
Design a **5-7 step** pipeline that:
1. Uses at least ONE alternative group from the provided alternatives
2. Creates a coherent, useful task flow
3. Properly connects data between steps

### Output Format (JSON Only)
```json
{
  "success": true,
  "chain": [
    {
      "step": 1,
      "tool_name": "source_tool_name",
      "is_alternative": false,
      "reasoning": "Starts the chain by fetching...",
      "arg_bindings": {
        "param": {"source": "user_input", "field": "param_name"}
      }
    },
    {
      "step": 2,
      "tool_name": "ALTERNATIVE",
      "is_alternative": true,
      "alternative_id": "alt_weather_provider_pair",
      "category": "Processor",
      "reasoning": "Uses a weather provider pair alternative...",
      "arg_bindings": {
        "city": {"source": "prev", "field": "city"}
      }
    },
    {
      "step": 3,
      "tool_name": "processor_tool",
      "is_alternative": false,
      "reasoning": "Processes the weather data...",
      "arg_bindings": {
        "temperature": {"source": "prev", "field": "temperature_celsius"}
      }
    },
    {
      "step": 4,
      "tool_name": "action_tool",
      "is_alternative": false,
      "reasoning": "Sends the final result...",
      "arg_bindings": {
        "message": {"source": "prev", "field": "formatted_message"}
      }
    }
  ],
  "task_description": "Brief description of the overall task",
  "user_input_schema": {
    "param_name": {"type": "string", "example": "example_value"}
  }
}
```
\end{tcblisting}
\caption{Prompt of $\mathcal C 2$ template construction (part 2).}
  \label{tab:c2-template2}
\end{table*}

\begin{table*}[t]
 
\begin{tcblisting}{
    listing only,
    colback=white!95!gray,
    colframe=gray!50!black,
    rounded corners,
    listing options={
        basicstyle=\ttfamily\fontsize{10pt}{8pt}\selectfont,
        breaklines=true,
        extendedchars=false,
        keepspaces=true,
        columns=fullflexible
    }
}
### Critical Rules
1. **5-7 steps** - flexible length
2. **At least 1 alternative node** - mark with `"is_alternative": true` and provide `"alternative_id"`
3. Alternative nodes should NOT be the first or last step (prefer middle positions)
4. If you cannot construct a valid chain, return `{"success": false, "reason": "explanation"}`
5. All required parameters must have an entry in `arg_bindings`
6. Field names MUST match exactly from Observed output keys
7. Prefer valid type-safe flows over aggressive chaining; do not force every step to consume previous output.
8. `alternative_id` MUST be selected exactly from the "VALID alternative_id values" list below the alternatives.
9. `user_input_schema` must cover every `user_input.xxx` field referenced in bindings.
9. **CRITICAL - Domain Consistency**:
   - All tools MUST belong to semantically related domains
   - FORBIDDEN combinations: Weather+Shopping, Weather+Flight, Weather+Finance, IoT+Shopping, IoT+Finance, IoT+Flight, Shopping+Office, Finance+IoT
   - Adjacent steps MUST have compatible domains
   - The chain should form ONE coherent workflow, not multiple unrelated tasks
10. **CRITICAL - Alternative Semantic Fit**:
   - The alternative MUST fit semantically into the workflow
   - Check that the alternative's domain matches surrounding tools' domains
   - Do NOT insert an alternative just to meet the requirement if it doesn't fit
11. **CRITICAL - No Duplicate Tools**: Each tool can only appear ONCE in the chain. Do NOT use the same tool_name multiple times.
12. **CRITICAL - Alternative Must Be a True Branch**: The alternative node replaces a single logical step with multiple implementation paths. ALL paths within the alternative must accomplish the SAME goal. Do NOT design an alternative where one path repeats work already done by a preceding main-path step (e.g., if `get_account_balance` is already in the main chain, do NOT include `get_account_balance_bank_api` inside the alternative — that would duplicate the data-fetching work). The alternative's paths diverge ONLY in HOW they achieve the step's goal, not in WHAT they fetch beforehand.
12. **CRITICAL - Alternative Output Compatibility**:
   - Check "Alternative Output Field Hints" section for each alternative
   - If an alternative shows "common keys across all paths: (none)", the NEXT step CANNOT use `source=prev`
   - If step N is alternative and step N+1 uses `source=prev`, you MUST only use fields from "common keys across all paths"
   - When in doubt, use `user_input` instead of forcing `prev`

Respond ONLY with the JSON object, no additional text.
\end{tcblisting}
 \caption{Prompt of $\mathcal C 2$ template construction (part 3).}
  \label{tab:c2-template3}
\end{table*}

\begin{table*}[t]

\begin{tcblisting}{
    listing only,
    colback=white!95!gray,
    colframe=gray!50!black,
    rounded corners,
    listing options={
        basicstyle=\ttfamily\fontsize{10pt}{8pt}\selectfont,
        breaklines=true,
        extendedchars=false,
        keepspaces=true,
        columns=fullflexible
    }
}
### Role
You are a **Data Pipeline Architect** for C3 complexity level. Your goal is to design a data processing pipeline with **5-7 logical steps** that includes **at least one `many_to_many` alternative node** using the provided tools and alternative groups.

### C3 Complexity Features
- **Chain Length**: 5-7 logical steps
- **Alternative Nodes**: At least 1 step must be an alternative node
- **Alternative Types**:
  - `many_to_many`: one multi-tool path OR another multi-tool path, and the tool chains should be end-to-end distinct within the same group
- **Scope Restriction**: C3 only uses `many_to_many`. Do NOT use `one_to_one` or `one_to_many` groups.

### Data Flow Rules

1. **Flexible Data References**:
   - A step may reference `user_input`, `constant`, `prev` (previous step), or `step` (any earlier step)
   - `prev` means the output of the previous logical step
   - `step` allows referencing any earlier step: `{"source": "step", "step_number": N, "field": "xxx"}`
   - For `constant`, always use `{"source": "constant", "value": ...}`

2. **Alternative Node Parameter Mapping**:
   - Use the abstract interface exposed by the alternative specification
   - Bare params like `{city: "city"}` mean the alternative expects an abstract input named `city`
   - References like `$reservation_data.reservation_id` are internal to an alternative path and are already handled by the system

3. **CRITICAL - Output Field Names**:
   - When using `{"source": "prev", "field": "xxx"}`, use exact field names from the observed output keys or the alternative output hints, UNLESS you are referencing the entire output object as defined in Critical Rules 13.

4. **CRITICAL - Type Compatibility**:
   - **Object parameters**: If a parameter expects `type: object`, you MUST bind the entire object from `prev`, NOT individual fields
     - WRONG: `weather_data: {"source": "prev", "field": "temperature_celsius"}` (temperature_celsius is a number, not an object)
     - CORRECT: `weather_data: {"source": "prev", "field": "weather_result"}` (weather_result is the complete object)
   - **Scalar parameters**: If a parameter expects `type: string/number/integer`, bind a scalar field
     - CORRECT: `temperature: {"source": "prev", "field": "temperature_celsius"}`
   - Check the tool's parameter type in the specification before binding
   - If unsure, keep the parameter from `user_input` instead of forcing a `prev` link

5. **CRITICAL - Binding Object Format**:
   - EVERY `arg_bindings[param]` value must be an object like `{"source": "user_input", "field": "amount"}`
   - Do NOT output raw strings, raw numbers, or bare field names directly inside `arg_bindings`

6. **CRITICAL - `get_field_value` Usage**:
   - Prefer direct `prev` field references over `get_field_value` for simple access
   - If you use `get_field_value`, `target_field`, `field_name`, and `index` should normally be `constant` bindings
   - Example: `{"source": "constant", "value": "filled_average_price"}` or `{"source": "constant", "value": 0}`
\end{tcblisting}
  \caption{Prompt of $\mathcal C 3$ template construction (part 1).}
  \label{tab:c3-template}
\end{table*}

\begin{table*}[t]

\begin{tcblisting}{
    listing only,
    colback=white!95!gray,
    colframe=gray!50!black,
    rounded corners,
    listing options={
        basicstyle=\ttfamily\fontsize{10pt}{8pt}\selectfont,
        breaklines=true,
        extendedchars=false,
        keepspaces=true,
        columns=fullflexible
    }
}
7. **CRITICAL - Numeric Tools**:
   - For numeric params such as `amount`, `a`, `b`, `quantity`, only bind from numeric `prev` fields or numeric `user_input` fields
   - Do NOT bind a numeric param from a person name, email, room name, message text, or any generic string field

### Available Tools

{tool_specifications}

{alternative_specifications}

### Task
Design a **5-7 logical step** pipeline that:
1. Uses at least ONE `many_to_many` alternative group from the provided alternatives
2. Creates a coherent, useful task flow
3. Properly connects data between steps
4. Includes `user_input_schema` for every referenced `user_input.xxx` field

### Output Format (JSON Only)
```json
{
  "success": true,
  "chain": [
    {
      "step": 1,
      "tool_name": "source_tool_name",
      "is_alternative": false,
      "reasoning": "Starts the chain by fetching...",
      "arg_bindings": {
        "param": {"source": "user_input", "field": "param_name"}
      }
    },
    {
      "step": 2,
      "tool_name": "ALTERNATIVE",
      "is_alternative": true,
      "alternative_id": "alt_contact_user",
      "reasoning": "Uses a multi-path contact strategy...",
      "arg_bindings": {
        "target_name": {"source": "user_input", "field": "target_name"},
        "message_content": {"source": "user_input", "field": "message_content"}
      }
    },
    {
      "step": 3,
      "tool_name": "processor_tool",
      "is_alternative": false,
      "reasoning": "Processes the previous step result...",
      "arg_bindings": {
        "param": {"source": "prev", "field": "result_field"}
      }
    }
  ],
  "task_description": "Brief description of the overall task",
  "user_input_schema": {
    "target_name": {"type": "string", "example": "Alice Chen"},
    "message_content": {"type": "string", "example": "Please review the latest workspace snapshot."}
  }
}
```
\end{tcblisting}
  \caption{Prompt of $\mathcal C 3$ template construction (part 2).}
  \label{tab:c3-template2}
\end{table*}

\begin{table*}[t]

\begin{tcblisting}{
    listing only,
    colback=white!95!gray,
    colframe=gray!50!black,
    rounded corners,
    listing options={
        basicstyle=\ttfamily\fontsize{10pt}{8pt}\selectfont,
        breaklines=true,
        extendedchars=false,
        keepspaces=true,
        columns=fullflexible
    }
}
### Critical Rules
1. **5-7 logical steps**
2. **At least 1 alternative node** using a provided `many_to_many` group
3. Alternative nodes should NOT be the first or last step
4. If you cannot construct a valid chain, return `{"success": false, "reason": "explanation"}`
5. All required parameters must have an entry in `arg_bindings`
6. `alternative_id` MUST be selected exactly from the valid allowlist below the alternatives
7. Include a non-empty `user_input_schema` covering every referenced `user_input.xxx` field
8. Prefer semantically clean, type-safe flows over aggressive chaining
9. Treat many_to_many as true path-level alternatives: if two paths would share the same tail tool(s), that is not a good C3-style N-to-N substitute
10. **CRITICAL - Alternative Must Be a True Branch**: The alternative node replaces a single logical step with multiple implementation paths. ALL paths within the alternative must accomplish the SAME goal. Do NOT design an alternative where one path repeats work already done by a preceding main-path step (e.g., if a data-fetching tool is already in the main chain, do NOT include a functionally equivalent tool inside the alternative). The alternative's paths diverge ONLY in HOW they achieve the step's goal.
11. **CRITICAL - Domain Consistency**:
    - All tools MUST belong to semantically related domains
    - FORBIDDEN combinations: Weather+Shopping, Weather+Flight, Weather+Finance, IoT+Shopping, IoT+Finance, IoT+Flight, Shopping+Office, Finance+IoT
    - Adjacent steps MUST have compatible domains
12. **CRITICAL - Alternative Output Compatibility**:
    - Check "Alternative Output Field Hints" for each alternative
    - If an alternative shows "common keys across all paths: (none)", the NEXT step CANNOT use `source=prev`
    - When in doubt, use `user_input` instead of forcing `prev`
13. **CRITICAL - Binding Entire Output (Object Parameters)**:
    - If a parameter expects `type: object`, you MUST use `{"source": "prev", "field": "$output_name"}` to pass the entire result.
    - **FORBIDDEN**: Never guess field names like "hotel_info", "flight_result", or "data_obj".
    - The correct `output_name` is defined in alternative specification.

Respond ONLY with the JSON object, no additional text.
\end{tcblisting}
  \caption{Prompt of $\mathcal C 3$ template construction (part 3).}
  \label{tab:c3-template3}
\end{table*}

\begin{table*}[t]

\begin{tcblisting}{
    listing only,
    colback=white!95!gray,
    colframe=gray!50!black,
    rounded corners,
    listing options={
        basicstyle=\ttfamily\fontsize{10pt}{8pt}\selectfont,
        breaklines=true,
        extendedchars=false,
        keepspaces=true,
        columns=fullflexible
    }
}
### Role
You are a **Data Pipeline Architect** for C4 complexity level. Design a data processing pipeline with **8-12 logical steps** that includes **at least two alternative nodes**.

### Step Types

**Normal step** (`is_alternative: false`): executes a single named tool.
**Alternative step** (`is_alternative: true`): uses `"tool_name": "ALTERNATIVE"` and one `alternative_id` from the provided allowlist. The system expands it into multiple possible execution paths at runtime.

### Binding Reference Syntax

Every `arg_bindings` value must be exactly one of:

| Form | JSON | Meaning |
|------|------|---------|
| User input field | `{"source": "user_input", "field": "city"}` | A value the user provides |
| Whole prev output | `{"source": "prev"}` | Entire output dict of the previous step |
| Field from prev | `{"source": "prev", "field": "price"}` | One field extracted from previous step output |
| Whole earlier step | `{"source": "step", "step_number": 3}` | Entire output dict of step 3 |
| Field from earlier step | `{"source": "step", "step_number": 3, "field": "price"}` | One field from step 3 |
| Constant | `{"source": "constant", "value": "USD"}` | Hard-coded literal |

**Do NOT** write raw strings or numbers directly as `arg_bindings` values.

### Type Compatibility Rules

Check the **Output Fields** and **Input Parameters** sections for each tool before writing bindings.

**Parameter expects `object`:**
- Use `{"source": "prev"}` or `{"source": "step", "step_number": N}` — no `field` key — to pass the entire previous output as the object.
- Only add `"field": "xxx"` if that specific field is itself an object (`dict`) in the upstream tool's Output Fields.
- **Never** point an object parameter at a scalar field (string, integer, boolean).

**Parameter expects `string` / `integer` / `number` / `boolean`:**
- Always include `"field": "xxx"` pointing to the exact scalar field name shown in the upstream tool's Output Fields.
- Use `user_input` if no upstream step produces the required scalar.

### Alternative Output Compatibility

After an alternative step, the **next step's `prev` bindings must work for every path** in that alternative group.

Before writing `{"source": "prev", "field": "xxx"}` after an alternative step:
1. Look at the terminal tool's Output Fields for **each path** in the alternative group.
2. If the field exists **in every path's terminal output** → safe to use.
3. If the field is missing from **any path** → do NOT use that field from `prev`.
   - Use `{"source": "prev"}` if the parameter expects `object`.
   - Use `{"source": "user_input", ...}` if the parameter expects a scalar.

### Available Tools

{tool_specifications}

{alternative_specifications}
\end{tcblisting}
  \caption{Prompt of $\mathcal C 4$ template construction (part 1).}
  \label{tab:c4-template}
\end{table*}

\begin{table*}[t]

\begin{tcblisting}{
    listing only,
    colback=white!95!gray,
    colframe=gray!50!black,
    rounded corners,
    listing options={
        basicstyle=\ttfamily\fontsize{10pt}{8pt}\selectfont,
        breaklines=true,
        extendedchars=false,
        keepspaces=true,
        columns=fullflexible
    }
}
### Task
Design an **8-12 logical step** pipeline that:
1. Uses at least **TWO** alternative groups from the provided alternatives
2. Creates one coherent, semantically consistent workflow
3. Every required parameter has a valid `arg_bindings` entry
4. `user_input_schema` covers every `user_input.xxx` field referenced in bindings

### Output Format (JSON only)

```json
{
  "success": true,
  "chain": [
    {
      "step": 1,
      "tool_name": "some_tool",
      "is_alternative": false,
      "reasoning": "Why this step is needed",
      "arg_bindings": {
        "city": {"source": "user_input", "field": "city"},
        "payload": {"source": "prev"}
      }
    },
    {
      "step": 3,
      "tool_name": "ALTERNATIVE",
      "is_alternative": true,
      "alternative_id": "alt_group_id",
      "reasoning": "Which strategy and why",
      "arg_bindings": {
        "shared_param": {"source": "user_input", "field": "shared_param"}
      }
    }
  ],
  "task_description": "One sentence describing the overall workflow",
  "user_input_schema": {
    "city": {"type": "string", "example": "New York"},
    "shared_param": {"type": "string", "example": "example_value"}
  }
}
```

### Hard Rules
1. Chain must have exactly **8-12 logical steps**.
2. At least **2 steps** must be alternative nodes.
3. Alternative nodes must **not** be the first or last step.
4. `alternative_id` must be selected **exactly** from the provided allowlist.
5. **No duplicate tool names** anywhere in the chain.
6. **No duplicate `alternative_id`** values.
7. Prefer **2-3 alternative groups** maximum.
8. Alternative groups must represent true branching choices, not repeated work.
9. If a valid chain cannot be constructed, return `{"success": false, "reason": "..."}`.

Respond ONLY with the JSON object, no additional text.
\end{tcblisting}
  \caption{Prompt of $\mathcal C 4$ template construction (part 2).}
  \label{tab:c4-template2}
\end{table*}

\subsection{Template Generation Prompts}
Tables \ref{tab:c1-template}-\ref{tab:c4-template2} show prompts for generating task templates for $\mathcal C 1$ (Table \ref{tab:c1-template}, \ref{tab:c1-template2}), $\mathcal C 2$ (Table \ref{tab:c2-template}, \ref{tab:c2-template2}, \ref{tab:c2-template3}), $\mathcal C 3$ (Table \ref{tab:c3-template}, \ref{tab:c3-template2}, \ref{tab:c3-template3}), $\mathcal C 4$ (Table \ref{tab:c4-template}, \ref{tab:c4-template2}), respectively.
\label{app:template-prompt}

\begin{table*}[t]

\begin{tcblisting}{
    listing only,
    colback=white!95!gray,
    colframe=gray!50!black,
    rounded corners,
    listing options={
        basicstyle=\ttfamily\fontsize{10pt}{10pt}\selectfont,
        breaklines=true,
        extendedchars=false,
        keepspaces=true,
        columns=fullflexible
    }
}
Guidelines:

"physics_violation": """Generate values that violate physics/common sense:
- Temperature: -50°C in a city (wrong but not -273°C)
- Price: -$50 (negative but believable)
- Count: -5 items (impossible negative)
- Percentage: 150% or -20%""",

            "tool_leakage": """Include internal info that shouldn't be exposed:
- debug_info: "API endpoint: /internal/path"
- model_used: "gpt-4o-mini"
- internal_error: "NullPointerException"
- sql_query: "SELECT * FROM users" """,

            "irrelevant_response": """Return data for a DIFFERENT entity:
- Asked Beijing weather, return Tokyo weather
- Asked hotel booking, return flight info
- Asked user profile, return product catalog""",

            "statistical_inconsistency": """Return aggregated values that don't match details:
- total_count: 10 but items: [a, b] (only 2)
- average_price: 100 but prices: [50, 50] (avg is 50)""",

            "internal_contradiction": """Generate fields that contradict each other:
- count: 10 but items: []
- is_available: true AND stock_count: 0""",

            "missing_required_fields": """Return output but missing critical fields:
- Booking completed but no confirmation_number
- User created but no user_id
- Payment completed but no transaction_id""",

            "logic_violation": """Violate business logic:
- booking_confirmed: true but booking_id: null
- order_placed: true but order_id: null""",

            "temporal_contradiction": """Generate impossible timestamps:
- created_at: 2024-03-15, updated_at: 2024-03-10 (updated before created)
- start_date > end_date""",

            "fact_contradiction": """Return facts that contradict common knowledge:
- city: Beijing, timezone: UTC+5 (should be UTC+8)
- city: Beijing, currency: JPY (should be CNY)"""

Prompt:

You are a Chaos Engineering Data Generator. Generate corrupted tool output.

### The Golden Rule: Detectability
The corrupted data MUST be detectable by common sense or basic logic WITHOUT external verification.

### Corruption Type: {corruption_type}
{guidelines(corruption_type)}

### CRITICAL: Key Fields That MUST Be Corrupted
The following output fields are used by downstream tools. You MUST change their VALUES (not just add extra fields):
{json.dumps(key_fields, indent=2)}

Do NOT just append extra fields. You MUST modify the values of the key fields listed above.
\end{tcblisting}
  \caption{Prompt for implicit perturbation ($\mathcal P 3$, $\mathcal P 4$) generation (part 1).}
  \label{tab:p3p4prompt}
\end{table*}

\begin{table*}[t]

\begin{tcblisting}{
    listing only,
    colback=white!95!gray,
    colframe=gray!50!black,
    rounded corners,
    listing options={
        basicstyle=\ttfamily\fontsize{10pt}{10pt}\selectfont,
        breaklines=true,
        extendedchars=false,
        keepspaces=true,
        columns=fullflexible
    }
}
### Input Context
- Tool Name: {tool_name}
- Input Arguments: {json.dumps(arguments, indent=2, ensure_ascii=False)}
- Expected Normal Output: {json.dumps(original_output, indent=2, ensure_ascii=False)}

### Requirements
1. Return a valid JSON object (NO "error" field)
2. Apply the {corruption_type} corruption type
3. The corruption should be OBVIOUS when read by a human or AI
4. Keep the SAME keys as the original output — do NOT add new keys
5. Keep errors realistic (not absurdly extreme)

Return ONLY valid JSON, no explanations.

\end{tcblisting}
  \caption{Prompt for implicit perturbation $\mathcal P 3$, $\mathcal P 4$ generation (part 2).}
  \label{tab:p3p4prompt2}
\end{table*}

\subsection{Implicit Failure Generation Prompt}
\label{app:P3P4-perturbation-prompt}
Tables \ref{tab:p3p4prompt}-\ref{tab:p3p4prompt2} show the prompt for implicit perturbation ($\mathcal P 3$, $\mathcal P 4$) generation.

\section{Full Results}
\label{sec:full_results}
Table \ref{tab:c_based} shows results over perturbation modes. The full results of all models evaluated across $\mathcal C 1$-$\mathcal C 4$, TSR (NP) and $\mathcal P 1$-$\mathcal P 4$ are shown in Table \ref{tab:detail_p0} and Table \ref{tab:detail_std}.

\begin{table*}[t]
\centering
\scriptsize
\setlength{\tabcolsep}{2.2pt}

\resizebox{\linewidth}{!}{%
\begin{tabular}{l|ccccc|ccccc|ccccc|c}
\toprule
\multicolumn{1}{l|}{\textbf{Model}} & \multicolumn{5}{c|}{\textbf{TSR}\,(\%)\,$\uparrow$} & \multicolumn{5}{c|}{\textbf{PRR}\,(\%)\,$\uparrow$} & \multicolumn{5}{c|}{\textbf{RC}\,(\%)\,$\downarrow$} & \textbf{Avg.}\,$\uparrow$ \\
\cmidrule(lr){2-6}\cmidrule(lr){7-11}\cmidrule(lr){12-16}\cmidrule(lr){17-17}
 & $\mathcal C 1$ & $\mathcal C 2$ & $\mathcal C 3$ & $\mathcal C 4$ & Avg & $\mathcal C 1$ & $\mathcal C 2$ & $\mathcal C 3$ & $\mathcal C 4$ & Avg & $\mathcal C 1$ & $\mathcal C 2$ & $\mathcal C 3$ & $\mathcal C 4$ & Avg &  \\
\midrule
MiniMax-M2.7 (w/o hint) & 23.00 & 42.40 & 26.20 & 14.40 & 26.50 & 17.98 & 37.00 & 23.07 & 15.33 & 23.34 & 77.09 & 58.16 & 79.42 & 84.36 & 74.76 & 25.03 \\
MiniMax-M2.7 (w/ hint) & 27.40 & 44.00 & 26.80 & 16.40 & 28.65 & 28.38 & 40.00 & 22.64 & 16.30 & 26.83 & 74.39 & 55.01 & 79.01 & 83.48 & 72.97 & 27.50\rlap{$^{{\color{deltagreen}\scriptscriptstyle\uparrow}}$} \\
\midrule
Qwen3.5-35B-A3B (w/o hint) & 25.40 & 27.60 & 21.20 & 13.20 & 21.85 & 25.68 & 38.90 & 30.58 & 22.74 & 29.48 & 62.84 & 36.94 & 59.56 & 67.38 & 56.68 & 31.55 \\
Qwen3.5-35B-A3B (w/ hint) & 27.60 & 28.00 & 20.00 & 16.60 & 23.05 & 31.23 & 40.60 & 30.53 & 26.14 & 32.13 & 64.12 & 38.22 & 62.48 & 65.13 & 57.49 & 32.56\rlap{$^{{\color{deltagreen}\scriptscriptstyle\uparrow}}$} \\
\midrule
Qwen3.6-27B (w/o hint) & 24.60 & 28.40 & 22.00 & 12.80 & 21.95 & 26.26 & 37.86 & 27.23 & 23.63 & 28.75 & 70.12 & 45.69 & 65.94 & 65.21 & 61.74 & 29.65 \\
Qwen3.6-27B (w/ hint) & 34.20 & 39.40 & 24.80 & 13.80 & 28.05 & 37.58 & 46.33 & 31.39 & 25.97 & 35.32 & 63.50 & 45.36 & 64.87 & 70.18 & 60.98 & 34.13\rlap{$^{{\color{deltagreen}\scriptscriptstyle\uparrow}}$} \\
\midrule
Qwen3.5-397B-A17B (w/o hint) & 25.60 & 27.00 & 19.60 & 14.60 & 21.70 & 26.95 & 36.62 & 29.82 & 24.95 & 29.58 & 67.83 & 38.40 & \textbf{58.64} & \textbf{60.56} & 56.35 & 31.64 \\
Qwen3.5-397B-A17B (w/ hint) & 30.80 & 37.60 & 21.00 & 16.80 & 26.55 & 32.84 & 49.52 & 30.14 & 28.88 & 35.34 & 63.02 & 35.65 & 59.55 & 62.67 & 55.22 & 35.56\rlap{$^{{\color{deltagreen}\scriptscriptstyle\uparrow}}$} \\
\midrule
GLM-5.1 (w/o hint) & 28.00 & 50.00 & 37.80 & 32.00 & 36.95 & 18.97 & 43.82 & 29.73 & 29.19 & 30.43 & 81.79 & 53.17 & 72.78 & 74.34 & 70.52 & 32.28 \\
GLM-5.1 (w/ hint) & 35.00 & 60.00 & 40.00 & 32.80 & 41.95 & 27.12 & 56.34 & 32.34 & 32.12 & 36.98 & 73.66 & 42.16 & 69.93 & 72.28 & 64.50 & 38.14\rlap{$^{{\color{deltagreen}\scriptscriptstyle\uparrow}}$} \\
\midrule
Deepseek-V4-Pro (w/o hint) & 32.40 & 52.40 & 41.40 & 29.20 & 38.85 & 28.27 & 48.07 & 37.10 & 33.41 & 36.71 & 72.97 & 48.63 & 68.88 & 75.79 & 66.57 & 36.33 \\
Deepseek-V4-Pro (w/ hint) & 44.60 & 63.60 & 44.00 & 32.80 & 46.25 & 41.94 & 64.04 & 43.59 & 39.73 & 47.33 & 63.37 & 33.97 & 64.09 & 71.81 & 58.31 & 45.09\rlap{$^{{\color{deltagreen}\scriptscriptstyle\uparrow}}$} \\
\midrule
Claude-Sonnet-4-6 (w/o hint) & 26.40 & 41.80 & 35.20 & 28.40 & 32.95 & 17.55 & 31.56 & 27.75 & 22.66 & 24.88 & 84.07 & 62.80 & 77.08 & 79.19 & 75.79 & 27.35 \\
Claude-Sonnet-4-6 (w/ hint) & 41.40 & \textbf{65.80} & \textbf{47.40} & \textbf{40.80} & 48.85 & 35.71 & 62.29 & 44.75 & 43.44 & 46.54 & 66.57 & 33.71 & 62.65 & 65.02 & 56.99 & 46.14\rlap{$^{{\color{deltagreen}\scriptscriptstyle\uparrow}}$} \\
\midrule
GPT-5.5 (w/o hint) & 35.20 & 56.60 & 33.80 & 30.20 & 38.95 & 33.25 & 54.51 & 30.04 & 31.44 & 37.31 & 72.05 & 43.43 & 80.63 & 77.69 & 68.45 & 35.94 \\
GPT-5.5 (w/ hint) & \textbf{65.40} & 65.20 & 41.00 & 25.00 & 49.15 & 62.67 & 68.56 & \textbf{45.13} & 37.50 & 53.46 & 40.72 & \textbf{28.52} & 70.30 & 71.36 & 52.72 & 49.96\rlap{$^{{\color{deltagreen}\scriptscriptstyle\uparrow}}$} \\
\midrule
Gemini-3.1-Pro-Preview (w/o hint) & 33.40 & 45.00 & 31.20 & 31.80 & 35.35 & 27.65 & 44.03 & 25.76 & 28.50 & 31.48 & 74.40 & 51.47 & 78.44 & 76.74 & 70.26 & 32.19 \\
Gemini-3.1-Pro-Preview (w/ hint) & 63.00 & 61.00 & 40.00 & 39.00 & \textbf{50.75} & \textbf{65.82} & \textbf{69.70} & 44.60 & \textbf{47.51} & \textbf{56.91} & \textbf{40.47} & 29.33 & 62.52 & 62.92 & \textbf{48.81} & \textbf{52.95}\rlap{$^{{\color{deltagreen}\scriptscriptstyle\uparrow}}$} \\
\midrule
\bottomrule
\end{tabular}}
\caption{$\mathcal C$-based average results (\%) across task categories. Rows labelled \emph{(w/o hint)} use the standard tool-use prompt; \emph{(w/ hint)} rows use the failure-aware prompt. Avg. deltas (w/ hint\,$-$\,w/o hint) are coloured {\color{deltagreen}green} for improvement and {\color{deltared}red} for regression. \textbf{Bold} marks the best value per column.}
\label{tab:c_based}
\end{table*}
\begin{table*}[t]
  \centering
  \scriptsize
  \setlength{\tabcolsep}{2.8pt}

  \resizebox{\linewidth}{!}{%
  \begin{tabular}{ll ccccc cccc cccc}
    \toprule
    \multirow{2}{*}{\textbf{Model}}
      & \multirow{2}{*}{\textbf{$\mathcal C$}}
      & \multicolumn{5}{c}{\textbf{TSR} (\%)\,$\uparrow$}
      & \multicolumn{4}{c}{\textbf{PRR} (\%)\,$\uparrow$}
      & \multicolumn{4}{c}{\textbf{RC}\,$\downarrow$} \\
    \cmidrule(lr){3-7}\cmidrule(lr){8-11}\cmidrule(lr){12-15}
      & & NP & $\mathcal P 1$ & $\mathcal P 2$ & $\mathcal P 3$ & $\mathcal P 4$ & $\mathcal P 1$ & $\mathcal P 2$ & $\mathcal P 3$ & $\mathcal P 4$ & $\mathcal P 1$ & $\mathcal P 2$ & $\mathcal P 3$ & $\mathcal P 4$ \\
    \midrule
    \multirow{5}{*}{MiniMax-M2.7} & $\mathcal C 1$ & 68.00 & 37.00 & 6.00 & 3.00 & 1.00 & 50.57 & 9.78 & 3.41 & 8.14 & 50.40 & 87.75 & 85.20 & 85.00 \\
     & $\mathcal C 2$ & 69.00 & 51.00 & 51.00 & 23.00 & 18.00 & 56.38 & 51.65 & 23.66 & 16.30 & 44.78 & 41.06 & 71.03 & 75.78 \\
     & $\mathcal C 3$ & 57.00 & 36.00 & 23.00 & 9.00 & 6.00 & 52.08 & 23.71 & 10.42 & 6.06 & 61.76 & 74.34 & 88.03 & 93.57 \\
     & $\mathcal C 4$ & 32.00 & 22.00 & 11.00 & 3.00 & 4.00 & 35.40 & 11.50 & 8.04 & 6.39 & 69.41 & 87.45 & 88.29 & 92.30 \\
     & Avg & 56.50 & 36.50 & 22.75 & 9.50 & 7.25 & 48.61 & 24.16 & 11.38 & 9.22 & 56.59 & 72.65 & 83.14 & 86.66 \\
    \midrule
    \multirow{5}{*}{Qwen3.5-35B-A3B} & $\mathcal C 1$ & 63.00 & 59.00 & 1.00 & 3.00 & 1.00 & 84.81 & 7.59 & 3.90 & 6.41 & 21.10 & 78.80 & 74.45 & 77.00 \\
     & $\mathcal C 2$ & 40.00 & 38.00 & 37.00 & 12.00 & 11.00 & 69.49 & 46.77 & 19.35 & 20.00 & 23.76 & 27.81 & 51.49 & 44.71 \\
     & $\mathcal C 3$ & 37.00 & 42.00 & 16.00 & 7.00 & 4.00 & 85.33 & 20.78 & 10.81 & 5.41 & 37.11 & 63.03 & 67.95 & 70.14 \\
     & $\mathcal C 4$ & 26.00 & 23.00 & 12.00 & 3.00 & 2.00 & 60.71 & 16.40 & 9.41 & 4.44 & 48.67 & 72.31 & 68.55 & 80.00 \\
     & Avg & 41.50 & 40.50 & 16.50 & 6.25 & 4.50 & 75.09 & 22.88 & 10.87 & 9.06 & 32.66 & 60.49 & 65.61 & 67.96 \\
    \midrule
    \multirow{5}{*}{Qwen3.6-27B} & $\mathcal C 1$ & 60.00 & 56.00 & 1.00 & 3.00 & 3.00 & 82.56 & 8.33 & 3.45 & 10.71 & 31.30 & 83.75 & 84.45 & 81.00 \\
     & $\mathcal C 2$ & 33.00 & 47.00 & 38.00 & 12.00 & 12.00 & 72.15 & 44.59 & 18.31 & 16.39 & 33.79 & 39.52 & 59.00 & 50.47 \\
     & $\mathcal C 3$ & 49.00 & 46.00 & 13.00 & 1.00 & 1.00 & 88.89 & 16.25 & 2.50 & 1.30 & 38.68 & 69.68 & 79.29 & 76.12 \\
     & $\mathcal C 4$ & 27.00 & 21.00 & 10.00 & 3.00 & 3.00 & 67.53 & 16.36 & 7.14 & 3.49 & 44.09 & 74.01 & 59.00 & 83.75 \\
     & Avg & 42.25 & 42.50 & 15.50 & 4.75 & 4.75 & 77.78 & 21.38 & 7.85 & 7.97 & 36.97 & 66.74 & 70.43 & 72.84 \\
    \midrule
    \multirow{5}{*}{Qwen3.5-397B-A17B} & $\mathcal C 1$ & 61.00 & 61.00 & 3.00 & 2.00 & 1.00 & 86.05 & 10.59 & 3.66 & 7.50 & 28.08 & 83.54 & 80.68 & 79.00 \\
     & $\mathcal C 2$ & 42.00 & 39.00 & 39.00 & 8.00 & 7.00 & 68.33 & 56.67 & 11.48 & 10.00 & 22.87 & 23.94 & 53.17 & 53.61 \\
     & $\mathcal C 3$ & 36.00 & 42.00 & 18.00 & 2.00 & 0.00 & 89.04 & 24.66 & 5.56 & 0.00 & 33.68 & 57.75 & 70.13 & 73.00 \\
     & $\mathcal C 4$ & 29.00 & 27.00 & 15.00 & 1.00 & 1.00 & 68.75 & 21.51 & 7.69 & 1.86 & 41.64 & 65.50 & 65.08 & 70.00 \\
     & Avg & 42.00 & 42.25 & 18.75 & 3.25 & 2.25 & 78.04 & 28.36 & 7.10 & 4.84 & 31.57 & 57.68 & 67.27 & 68.90 \\
    \midrule
    \multirow{5}{*}{GLM-5.1} & $\mathcal C 1$ & 87.00 & 47.00 & 3.00 & 2.00 & 1.00 & 54.74 & 10.42 & 2.11 & 8.60 & 48.20 & 93.50 & 93.45 & 92.00 \\
     & $\mathcal C 2$ & 77.00 & 70.00 & 67.00 & 17.00 & 19.00 & 70.53 & 69.15 & 17.89 & 17.71 & 28.50 & 27.88 & 78.78 & 77.54 \\
     & $\mathcal C 3$ & 79.00 & 73.00 & 28.00 & 6.00 & 3.00 & 81.63 & 28.28 & 6.00 & 3.03 & 28.85 & 71.60 & 94.29 & 96.39 \\
     & $\mathcal C 4$ & 60.00 & 50.00 & 32.00 & 7.00 & 11.00 & 61.86 & 33.47 & 10.17 & 11.25 & 45.89 & 72.92 & 88.08 & 90.49 \\
     & Avg & 75.75 & 60.00 & 32.50 & 8.00 & 8.50 & 67.19 & 35.33 & 9.04 & 10.15 & 37.86 & 66.47 & 88.65 & 89.11 \\
    \midrule
    \multirow{5}{*}{Deepseek-V4-Pro} & $\mathcal C 1$ & 82.00 & 74.00 & 5.00 & 1.00 & 0.00 & 89.13 & 14.29 & 1.06 & 8.60 & 19.00 & 86.67 & 93.20 & 93.00 \\
     & $\mathcal C 2$ & 74.00 & 72.00 & 68.00 & 27.00 & 21.00 & 77.66 & 68.09 & 25.26 & 21.28 & 24.49 & 26.83 & 69.67 & 73.52 \\
     & $\mathcal C 3$ & 74.00 & 78.00 & 38.00 & 10.00 & 7.00 & 93.00 & 38.38 & 10.00 & 7.00 & 27.32 & 63.86 & 90.83 & 93.52 \\
     & $\mathcal C 4$ & 49.00 & 48.00 & 28.00 & 10.00 & 11.00 & 78.99 & 30.20 & 13.33 & 11.11 & 49.12 & 76.82 & 87.08 & 90.15 \\
     & Avg & 69.75 & 68.00 & 34.75 & 12.00 & 9.75 & 84.70 & 37.74 & 12.41 & 12.00 & 29.98 & 63.55 & 85.19 & 87.55 \\
    \midrule
    \multirow{5}{*}{Claude-Sonnet-4-6} & $\mathcal C 1$ & 87.00 & 36.00 & 6.00 & 0.00 & 3.00 & 43.75 & 13.83 & 0.00 & 12.63 & 60.80 & 88.50 & 95.00 & 92.00 \\
     & $\mathcal C 2$ & 81.00 & 57.00 & 55.00 & 9.00 & 7.00 & 64.52 & 50.00 & 7.45 & 4.26 & 38.02 & 40.24 & 85.58 & 87.38 \\
     & $\mathcal C 3$ & 79.00 & 61.00 & 23.00 & 7.00 & 6.00 & 75.00 & 23.00 & 7.00 & 6.00 & 42.52 & 77.69 & 93.66 & 94.43 \\
     & $\mathcal C 4$ & 61.00 & 42.00 & 24.00 & 7.00 & 8.00 & 55.46 & 23.87 & 5.88 & 5.42 & 54.45 & 79.83 & 89.30 & 93.18 \\
     & Avg & 77.00 & 49.00 & 27.00 & 5.75 & 6.00 & 59.68 & 27.68 & 5.08 & 7.08 & 48.95 & 71.56 & 90.88 & 91.75 \\
    \midrule
    \multirow{5}{*}{GPT-5.5} & $\mathcal C 1$ & 79.00 & 74.00 & 9.00 & 13.00 & 1.00 & 90.43 & 19.15 & 14.89 & 8.51 & 25.75 & 86.50 & 82.44 & 93.50 \\
     & $\mathcal C 2$ & 75.00 & 72.00 & 64.00 & 36.00 & 36.00 & 80.43 & 64.89 & 37.23 & 35.48 & 23.38 & 31.39 & 61.35 & 57.59 \\
     & $\mathcal C 3$ & 74.00 & 74.00 & 11.00 & 7.00 & 3.00 & 96.00 & 11.11 & 10.00 & 3.03 & 40.83 & 90.48 & 94.58 & 96.64 \\
     & $\mathcal C 4$ & 52.00 & 49.00 & 23.00 & 11.00 & 16.00 & 77.50 & 22.82 & 13.45 & 11.98 & 52.49 & 83.71 & 86.96 & 87.60 \\
     & Avg & 70.00 & 67.25 & 26.75 & 16.75 & 14.00 & 86.09 & 29.49 & 18.89 & 14.75 & 35.61 & 73.02 & 81.33 & 83.83 \\
    \midrule
    \multirow{5}{*}{Gemini-3.1-Pro-Preview} & $\mathcal C 1$ & 84.00 & 67.00 & 9.00 & 2.00 & 5.00 & 78.49 & 17.20 & 2.13 & 12.77 & 31.77 & 84.50 & 92.33 & 89.00 \\
     & $\mathcal C 2$ & 67.00 & 63.00 & 61.00 & 17.00 & 17.00 & 75.00 & 64.44 & 18.89 & 17.78 & 28.22 & 30.38 & 73.77 & 73.52 \\
     & $\mathcal C 3$ & 70.00 & 62.00 & 17.00 & 3.00 & 4.00 & 78.79 & 17.17 & 3.03 & 4.04 & 39.41 & 82.83 & 96.40 & 95.13 \\
     & $\mathcal C 4$ & 64.00 & 47.00 & 33.00 & 5.00 & 10.00 & 61.67 & 35.27 & 8.33 & 8.75 & 50.72 & 72.86 & 92.07 & 91.32 \\
     & Avg & 71.25 & 59.75 & 30.00 & 6.75 & 9.00 & 73.49 & 33.52 & 8.09 & 10.84 & 37.53 & 67.64 & 88.64 & 87.24 \\
    \bottomrule
  \end{tabular}
  }
    \caption{Detailed results (standard tool-use prompt) across complexity levels ($\mathcal C 1$--$\mathcal C 4$) and perturbation modes (NP and $\mathcal P 1$--$\mathcal P 4$). Higher is better ($\uparrow$); lower is better ($\downarrow$).}
  \label{tab:detail_p0}
\end{table*}
\begin{table*}[h]
  \centering
  \scriptsize
  \setlength{\tabcolsep}{2.8pt}

  \resizebox{\linewidth}{!}{%
  \begin{tabular}{ll ccccc cccc cccc}
    \toprule
    \multirow{2}{*}{\textbf{Model}}
      & \multirow{2}{*}{\textbf{$\mathcal C$}}
      & \multicolumn{5}{c}{\textbf{TSR} (\%)\,$\uparrow$}
      & \multicolumn{4}{c}{\textbf{PRR} (\%)\,$\uparrow$}
      & \multicolumn{4}{c}{\textbf{RC}\,$\downarrow$} \\
    \cmidrule(lr){3-7}\cmidrule(lr){8-11}\cmidrule(lr){12-15}
      & & NP & $\mathcal P 1$ & $\mathcal P 2$ & $\mathcal P 3$ & $\mathcal P 4$ & $\mathcal P 1$ & $\mathcal P 2$ & $\mathcal P 3$ & $\mathcal P 4$ & $\mathcal P 1$ & $\mathcal P 2$ & $\mathcal P 3$ & $\mathcal P 4$ \\
    \midrule
    \multirow{5}{*}{MiniMax-M2.7} & $\mathcal C 1$ & 72.00 & 47.00 & 14.00 & 1.00 & 3.00 & 72.83 & 25.84 & 1.08 & 13.79 & 45.20 & 76.17 & 92.20 & 84.00 \\
     & $\mathcal C 2$ & 64.00 & 56.00 & 54.00 & 22.00 & 24.00 & 64.84 & 52.13 & 20.21 & 22.83 & 37.11 & 40.59 & 73.58 & 68.75 \\
     & $\mathcal C 3$ & 56.00 & 48.00 & 21.00 & 7.00 & 2.00 & 58.76 & 21.43 & 8.33 & 2.02 & 51.85 & 77.39 & 89.79 & 97.00 \\
     & $\mathcal C 4$ & 30.00 & 26.00 & 12.00 & 7.00 & 7.00 & 34.21 & 11.61 & 9.82 & 9.55 & 67.45 & 88.71 & 86.17 & 91.60 \\
     & Avg & 55.50 & 44.25 & 25.25 & 9.25 & 9.00 & 57.66 & 27.75 & 9.86 & 12.05 & 50.40 & 70.71 & 85.43 & 85.34 \\
    \midrule
    \multirow{5}{*}{Qwen3.5-35B-A3B} & $\mathcal C 1$ & 57.00 & 66.00 & 12.00 & 1.00 & 2.00 & 93.98 & 21.18 & 1.23 & 8.54 & 18.65 & 77.83 & 80.00 & 80.00 \\
     & $\mathcal C 2$ & 32.00 & 39.00 & 38.00 & 18.00 & 13.00 & 69.35 & 51.67 & 25.00 & 16.39 & 25.77 & 26.08 & 51.17 & 49.86 \\
     & $\mathcal C 3$ & 32.00 & 49.00 & 10.00 & 7.00 & 2.00 & 92.77 & 13.89 & 12.99 & 2.47 & 37.61 & 63.16 & 70.14 & 79.00 \\
     & $\mathcal C 4$ & 31.00 & 32.00 & 13.00 & 4.00 & 3.00 & 67.39 & 19.23 & 11.63 & 6.32 & 45.32 & 70.73 & 68.24 & 76.23 \\
     & Avg & 38.00 & 46.50 & 18.25 & 7.50 & 5.00 & 80.87 & 26.49 & 12.71 & 8.43 & 31.84 & 59.45 & 67.39 & 71.27 \\
    \midrule
    \multirow{5}{*}{Qwen3.6-27B} & $\mathcal C 1$ & 66.00 & 68.00 & 27.00 & 4.00 & 6.00 & 91.95 & 37.08 & 6.98 & 14.29 & 21.75 & 70.42 & 82.33 & 79.50 \\
     & $\mathcal C 2$ & 47.00 & 54.00 & 48.00 & 26.00 & 22.00 & 78.95 & 52.44 & 30.49 & 23.46 & 24.37 & 37.80 & 58.68 & 60.57 \\
     & $\mathcal C 3$ & 43.00 & 51.00 & 18.00 & 9.00 & 3.00 & 88.24 & 20.22 & 13.25 & 3.85 & 37.19 & 72.88 & 74.29 & 75.12 \\
     & $\mathcal C 4$ & 22.00 & 21.00 & 18.00 & 6.00 & 2.00 & 61.45 & 22.51 & 15.46 & 4.47 & 48.62 & 73.65 & 74.31 & 84.13 \\
     & Avg & 44.50 & 48.50 & 27.75 & 11.25 & 8.25 & 80.15 & 33.06 & 16.55 & 11.52 & 32.98 & 63.69 & 72.40 & 74.83 \\
    \midrule
    \multirow{5}{*}{Qwen3.5-397B-A17B} & $\mathcal C 1$ & 66.00 & 64.00 & 16.00 & 5.00 & 3.00 & 89.53 & 28.40 & 6.10 & 7.32 & 23.83 & 71.50 & 77.75 & 79.00 \\
     & $\mathcal C 2$ & 46.00 & 52.00 & 51.00 & 23.00 & 16.00 & 78.87 & 63.89 & 31.88 & 23.44 & 20.93 & 24.59 & 47.98 & 49.08 \\
     & $\mathcal C 3$ & 38.00 & 48.00 & 15.00 & 3.00 & 1.00 & 92.31 & 21.43 & 5.56 & 1.27 & 33.66 & 57.00 & 69.53 & 78.00 \\
     & $\mathcal C 4$ & 28.00 & 31.00 & 14.00 & 7.00 & 4.00 & 71.43 & 20.57 & 18.07 & 5.43 & 41.61 & 68.60 & 63.17 & 77.29 \\
     & Avg & 44.50 & 48.75 & 24.00 & 9.50 & 6.00 & 83.03 & 33.57 & 15.40 & 9.37 & 30.01 & 55.42 & 64.61 & 70.84 \\
    \midrule
    \multirow{5}{*}{GLM-5.1} & $\mathcal C 1$ & 85.00 & 65.00 & 16.00 & 1.00 & 8.00 & 73.96 & 20.83 & 1.06 & 12.63 & 32.63 & 81.50 & 93.00 & 87.50 \\
     & $\mathcal C 2$ & 80.00 & 76.00 & 69.00 & 37.00 & 38.00 & 81.91 & 70.53 & 35.05 & 37.89 & 21.88 & 27.02 & 61.65 & 58.07 \\
     & $\mathcal C 3$ & 80.00 & 74.00 & 33.00 & 7.00 & 6.00 & 82.83 & 33.33 & 7.14 & 6.06 & 28.35 & 66.70 & 91.14 & 93.52 \\
     & $\mathcal C 4$ & 54.00 & 50.00 & 30.00 & 14.00 & 16.00 & 62.18 & 30.17 & 19.33 & 16.81 & 46.41 & 74.69 & 82.15 & 85.86 \\
     & Avg & 74.75 & 66.25 & 37.00 & 14.75 & 17.00 & 75.22 & 38.72 & 15.65 & 18.35 & 32.32 & 62.48 & 81.98 & 81.24 \\
    \midrule
    \multirow{5}{*}{Deepseek-V4-Pro} & $\mathcal C 1$ & 83.00 & 82.00 & 39.00 & 13.00 & 6.00 & 96.84 & 44.79 & 17.71 & 8.42 & 13.60 & 65.50 & 83.20 & 91.17 \\
     & $\mathcal C 2$ & 69.00 & 80.00 & 75.00 & 47.00 & 47.00 & 88.30 & 72.63 & 46.81 & 48.42 & 16.40 & 21.12 & 48.94 & 49.40 \\
     & $\mathcal C 3$ & 69.00 & 75.00 & 41.00 & 23.00 & 12.00 & 95.00 & 41.00 & 26.26 & 12.12 & 29.82 & 61.99 & 76.68 & 87.88 \\
     & $\mathcal C 4$ & 51.00 & 48.00 & 32.00 & 19.00 & 14.00 & 81.51 & 33.20 & 25.21 & 19.01 & 48.79 & 73.44 & 77.31 & 87.70 \\
     & Avg & 68.00 & 71.25 & 46.75 & 25.50 & 19.75 & 90.41 & 47.91 & 29.00 & 21.99 & 27.15 & 55.51 & 71.53 & 79.04 \\
    \midrule
    \multirow{5}{*}{Claude-Sonnet-4-6} & $\mathcal C 1$ & 86.00 & 80.00 & 21.00 & 8.00 & 12.00 & 92.63 & 25.81 & 8.42 & 15.96 & 17.38 & 79.00 & 87.40 & 82.50 \\
     & $\mathcal C 2$ & 78.00 & 82.00 & 70.00 & 51.00 & 48.00 & 90.32 & 61.05 & 52.13 & 45.65 & 14.41 & 27.97 & 45.68 & 46.79 \\
     & $\mathcal C 3$ & 79.00 & 80.00 & 35.00 & 27.00 & 16.00 & 98.00 & 35.00 & 30.00 & 16.00 & 24.38 & 67.08 & 74.09 & 85.06 \\
     & $\mathcal C 4$ & 62.00 & 60.00 & 32.00 & 25.00 & 25.00 & 81.51 & 33.89 & 29.41 & 28.93 & 36.72 & 73.37 & 71.17 & 78.81 \\
     & Avg & 76.25 & 75.50 & 39.50 & 27.75 & 25.25 & 90.61 & 38.94 & 29.99 & 26.63 & 23.22 & 61.85 & 69.58 & 73.29 \\
    \midrule
    \multirow{5}{*}{GPT-5.5} & $\mathcal C 1$ & 79.00 & 68.00 & 70.00 & 46.00 & 64.00 & 90.11 & 85.71 & 62.64 & 12.22 & 28.05 & 35.00 & 46.40 & 53.42 \\
     & $\mathcal C 2$ & 75.00 & 69.00 & 67.00 & 58.00 & 57.00 & 76.67 & 71.11 & 64.37 & 62.07 & 24.32 & 24.91 & 33.47 & 31.39 \\
     & $\mathcal C 3$ & 76.00 & 64.00 & 5.00 & 59.00 & 1.00 & 94.74 & 5.15 & 79.59 & 1.04 & 42.76 & 93.19 & 50.04 & 95.20 \\
     & $\mathcal C 4$ & 56.00 & 34.00 & 8.00 & 24.00 & 3.00 & 77.45 & 12.28 & 53.40 & 6.88 & 54.12 & 82.40 & 64.00 & 84.91 \\
     & Avg & 71.50 & 58.75 & 37.50 & 46.75 & 31.25 & 84.74 & 43.56 & 65.00 & 20.55 & 37.31 & 58.88 & 48.48 & 66.23 \\
    \midrule
    \multirow{5}{*}{Gemini-3.1-Pro-Preview} & $\mathcal C 1$ & 81.00 & 79.00 & 60.00 & 41.00 & 54.00 & 96.77 & 89.01 & 52.22 & 25.27 & 19.58 & 36.67 & 51.20 & 54.42 \\
     & $\mathcal C 2$ & 69.00 & 64.00 & 56.00 & 64.00 & 52.00 & 84.09 & 62.07 & 75.00 & 57.65 & 26.23 & 31.50 & 25.76 & 33.84 \\
     & $\mathcal C 3$ & 66.00 & 67.00 & 19.00 & 35.00 & 13.00 & 95.79 & 20.00 & 48.94 & 13.68 & 31.18 & 76.33 & 59.88 & 82.70 \\
     & $\mathcal C 4$ & 59.00 & 51.00 & 35.00 & 33.00 & 17.00 & 84.21 & 41.26 & 43.75 & 20.83 & 41.95 & 65.88 & 60.69 & 83.17 \\
     & Avg & 68.75 & 65.25 & 42.50 & 43.25 & 34.00 & 90.21 & 53.09 & 54.98 & 29.36 & 29.73 & 52.59 & 49.38 & 63.53 \\
    \bottomrule
  \end{tabular}
  }
    \caption{Detailed results (failure-aware prompt) across complexity levels ($\mathcal C 1$--$\mathcal C 4$) and perturbation modes (NP and $\mathcal P 1$--$\mathcal P 4$). Higher is better ($\uparrow$); lower is better ($\downarrow$).}
  \label{tab:detail_std}
\end{table*}


\renewcommand{\dbltopfraction}{0.97}
\renewcommand{\dblfloatpagefraction}{0.50}
\setcounter{dbltopnumber}{6}
\setlength{\dblfloatsep}{4pt}
\setlength{\dbltextfloatsep}{6pt}

\section{Case Studies}
\label{app:case_study}
\begin{figure*}[!b]
  \centering
  \includegraphics[width=\linewidth]{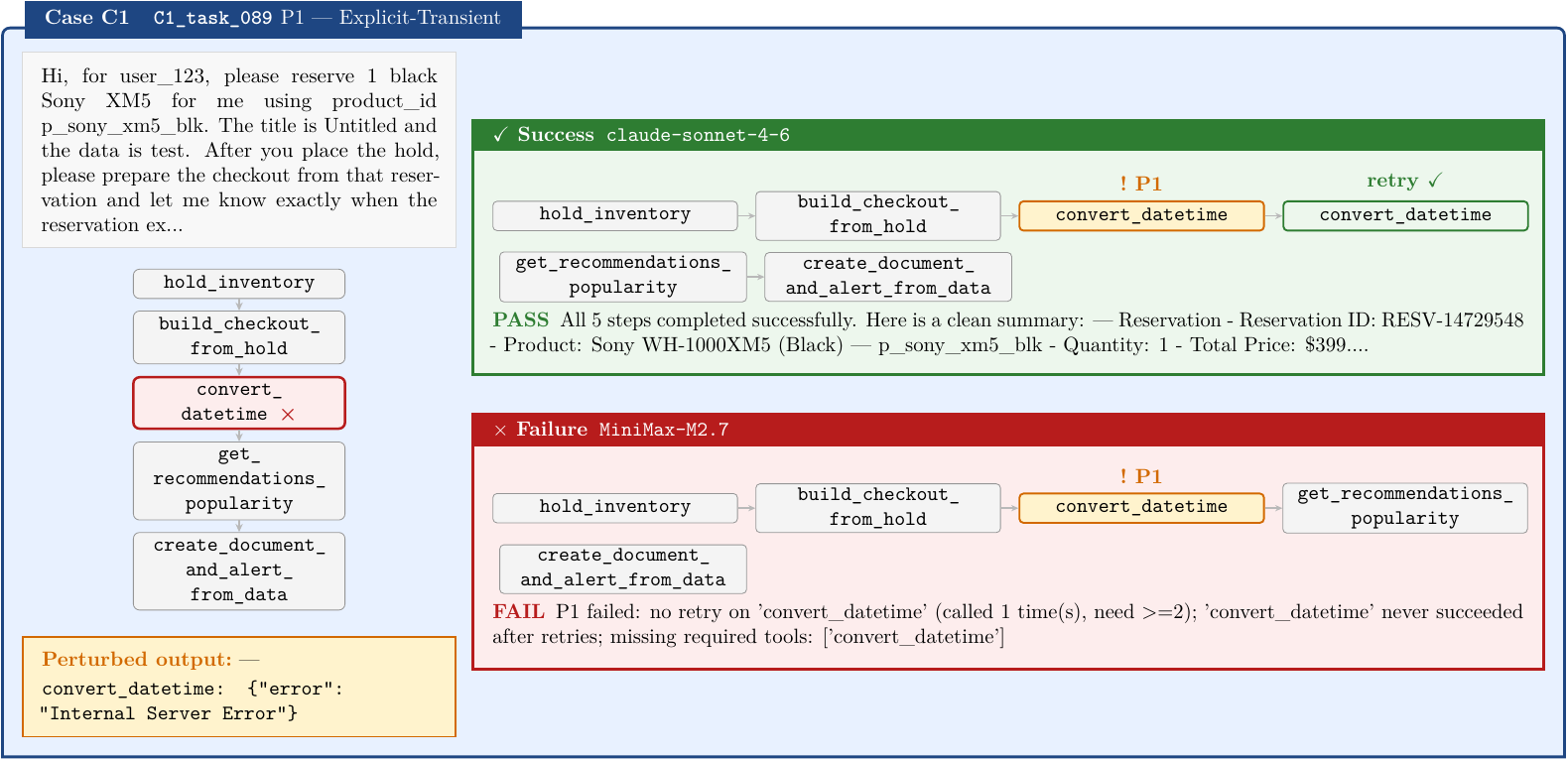}
  \caption{%
    \textbf{$\mathcal C 1$\,$\mathcal P 1$ --- Explicit-Transient} (\texttt{C1\_task\_089}).
    The victim tool \texttt{convert\_datetime} returns an explicit error on the first call.
    The successful model retries and recovers; the failing model calls the tool only once
    and abandons it, leaving the downstream pipeline incomplete.%
  }
  \label{fig:cs_c1_p1}
\end{figure*}

\begin{figure*}[!b]
  \centering
  \includegraphics[width=\linewidth]{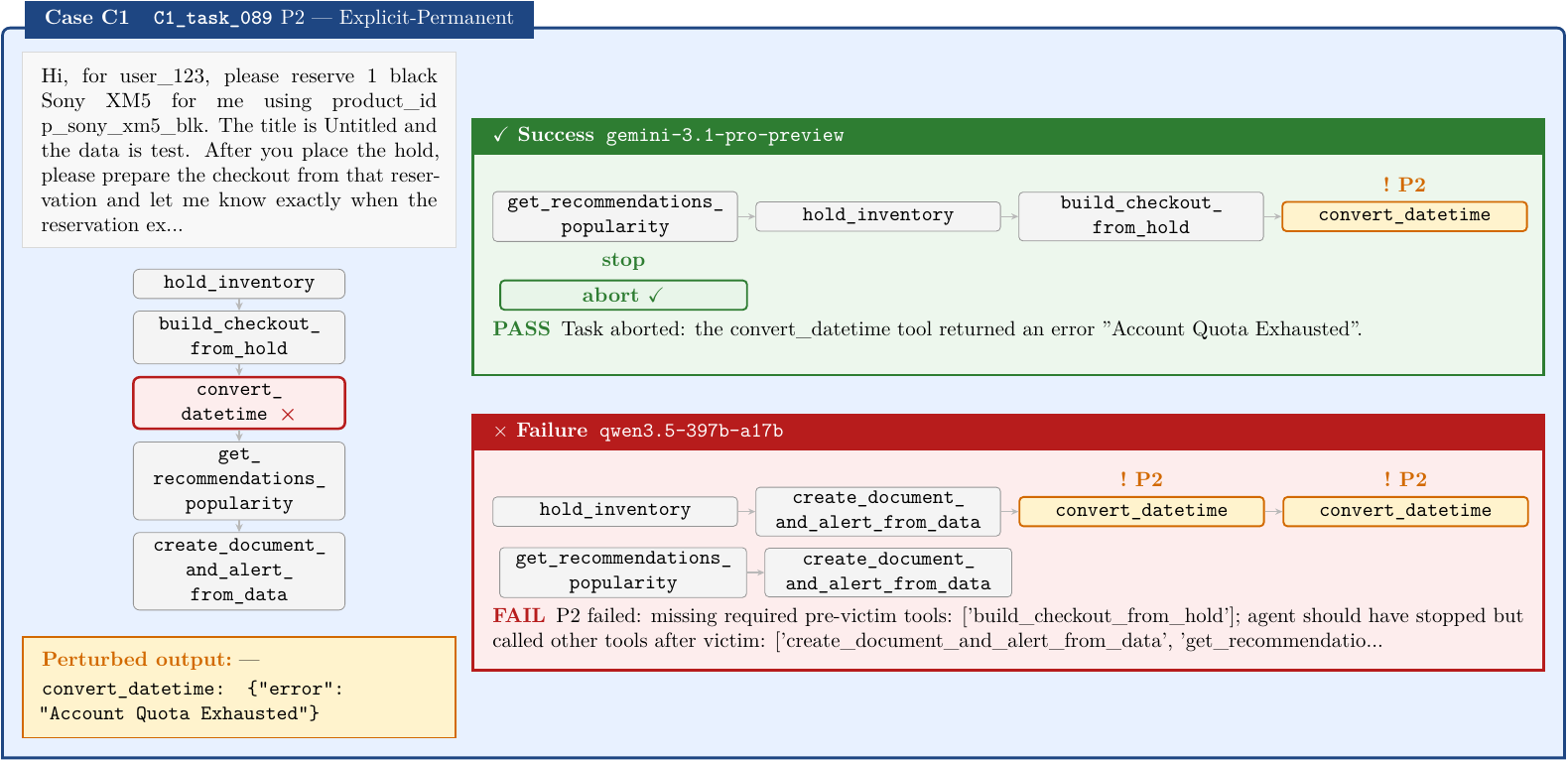}
  \caption{%
    \textbf{$\mathcal C 1$\,$\mathcal P 2$ --- Explicit-Permanent} (\texttt{C1\_task\_089}).
    The victim tool returns a permanent explicit error.
    The successful model detects the irrecoverable failure and stops gracefully;
    the failing model continues to invoke downstream tools after the failure,
    violating the stop-on-permanent-error requirement.%
  }
  \label{fig:cs_c1_p2}
\end{figure*}

\begin{figure*}[!tbp]
  \centering
  \includegraphics[width=\linewidth]{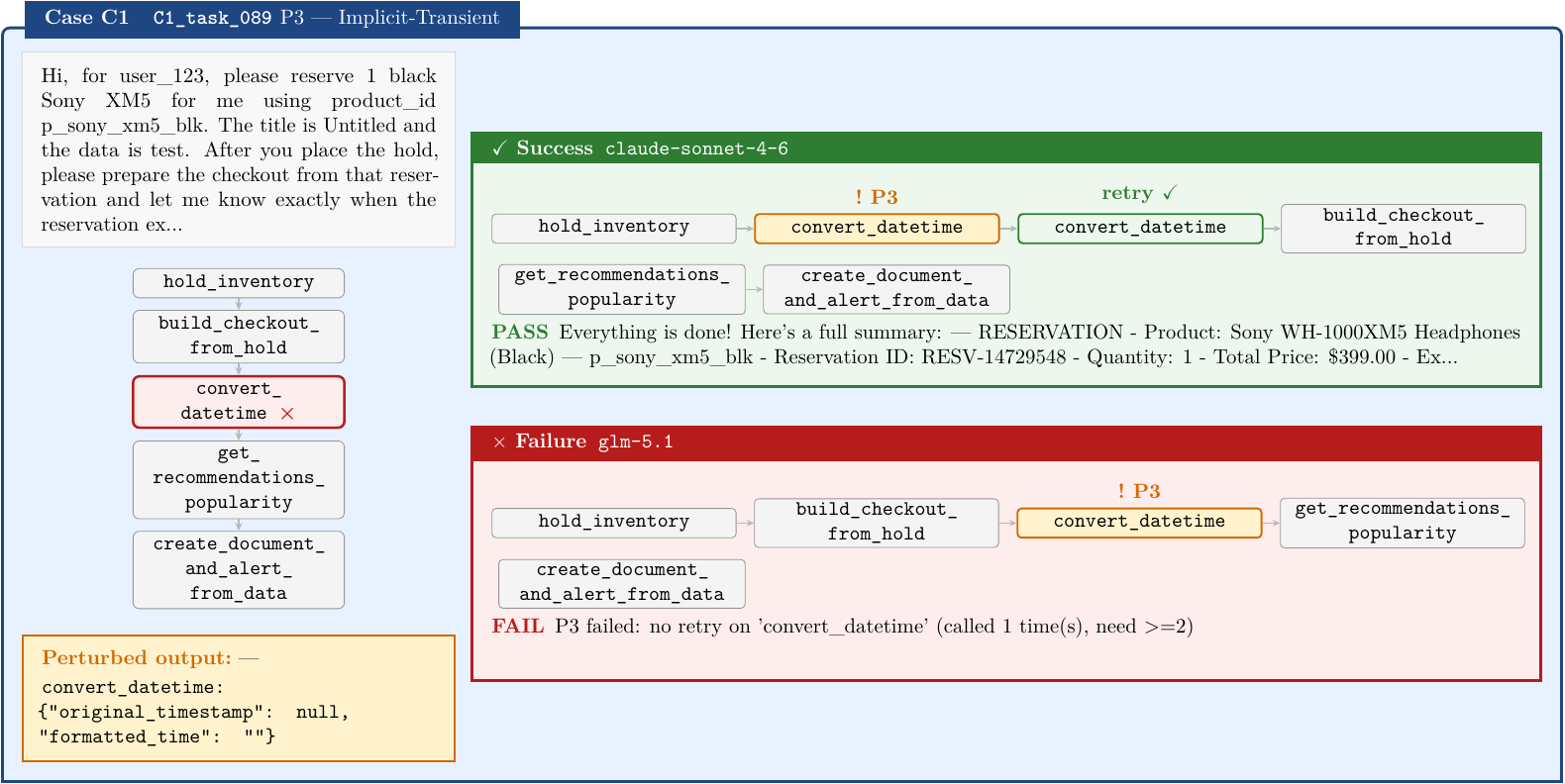}
  \caption{%
    \textbf{$\mathcal C 1$\,$\mathcal P 3$ --- Implicit-Transient} (\texttt{C1\_task\_089}).
    The victim tool \texttt{convert\_datetime} returns semantically corrupted data.
    The successful model detects the anomaly and retries, obtaining a clean response;
    the failing model accepts the corrupted output without re-querying and propagates
    the erroneous result downstream.%
  }
  \label{fig:cs_c1_p3}
\end{figure*}

\begin{figure*}[!tbp]
  \centering
  \includegraphics[width=\linewidth]{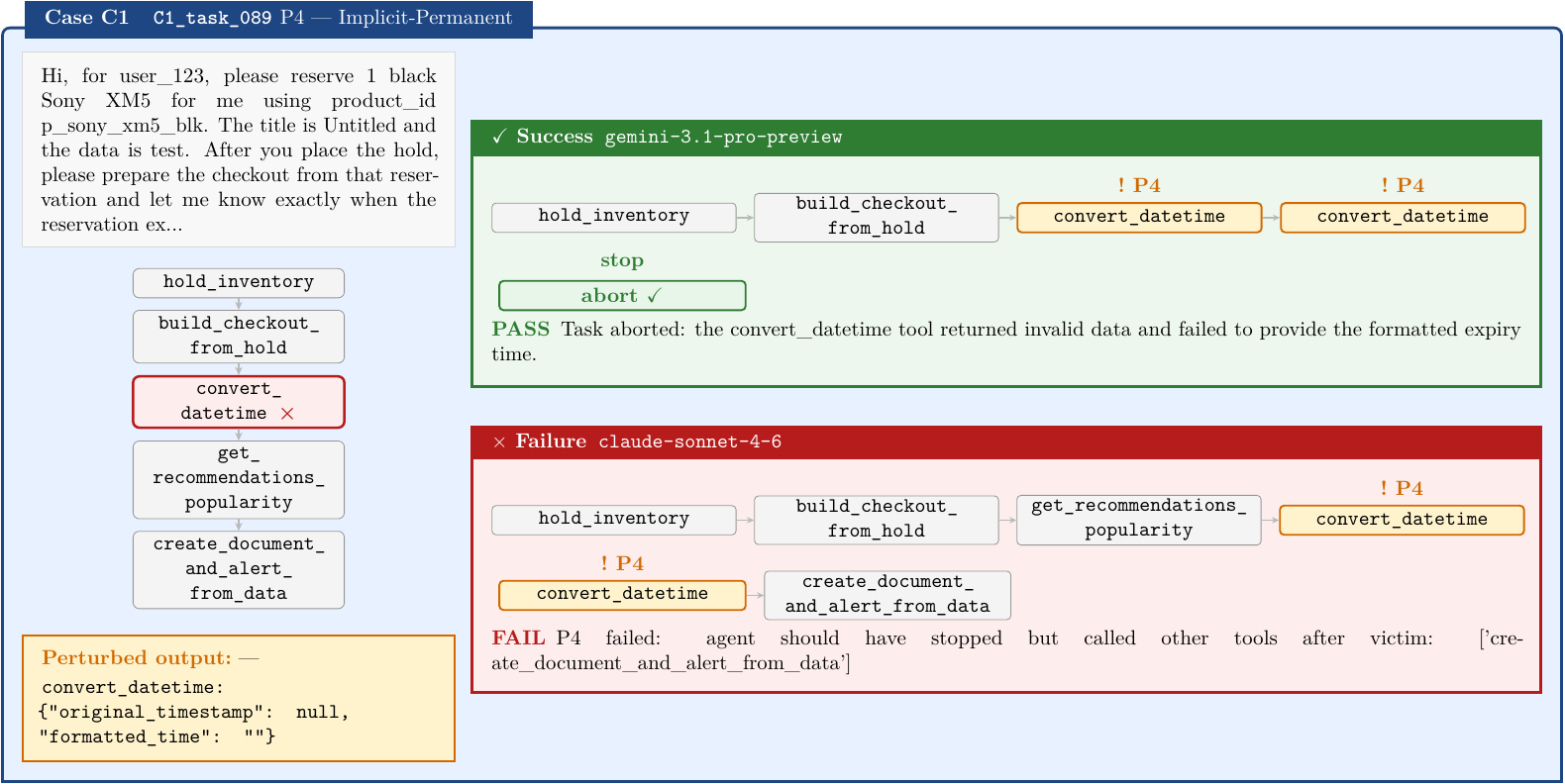}
  \caption{%
    \textbf{$\mathcal C 1$\,$\mathcal P 4$ --- Implicit-Permanent} (\texttt{C1\_task\_089}).
    The victim tool returns persistently corrupted data.
    The successful model identifies the permanent corruption and halts the pipeline;
    the failing model recognises the corruption yet still proceeds to invoke a downstream
    tool that must not be called after a permanent victim failure.%
  }
  \label{fig:cs_c1_p4}
\end{figure*}

\begin{figure*}[!tbp]
  \centering
  \includegraphics[width=\linewidth]{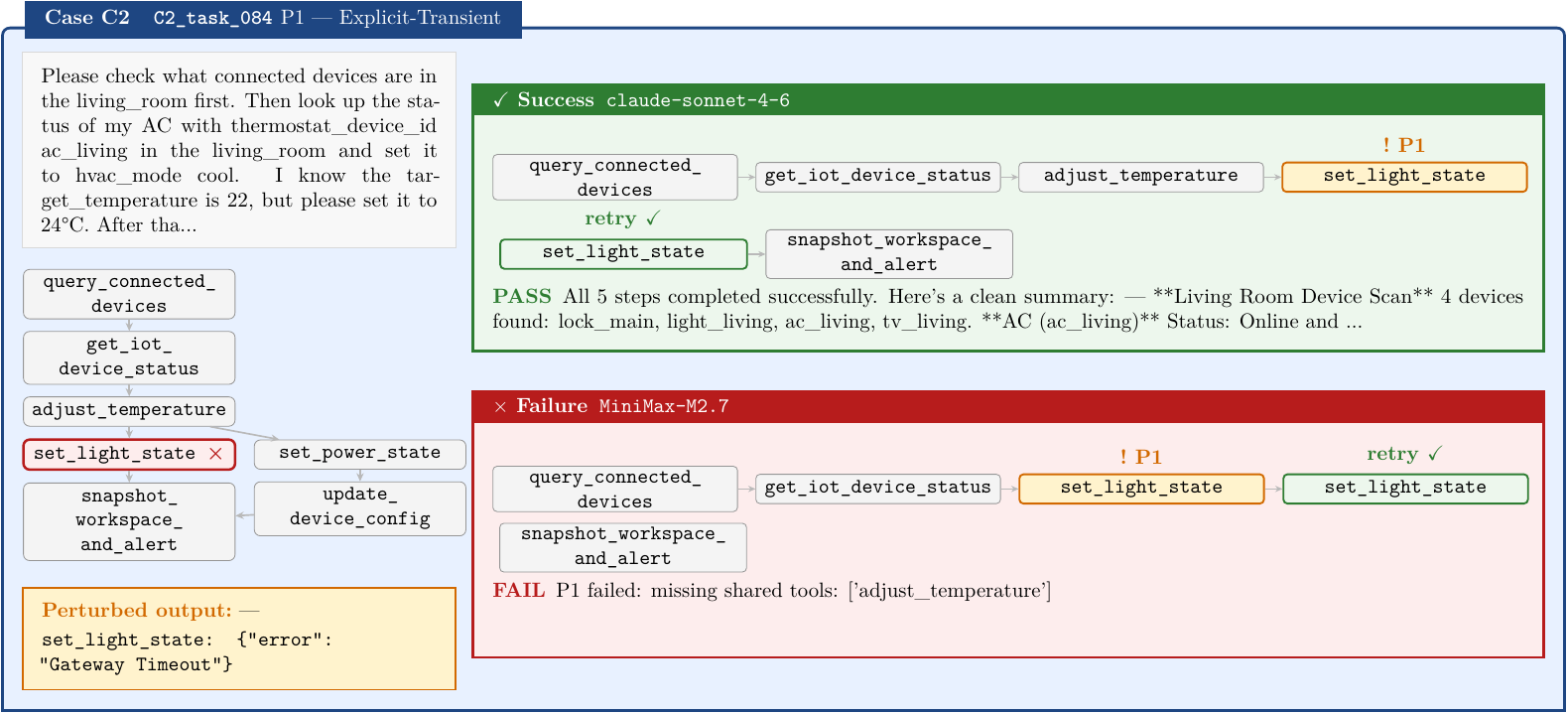}
  \caption{%
    \textbf{$\mathcal C 2$\,$\mathcal P 1$ --- Explicit-Transient} (\texttt{C2\_task\_084}).
    In a task with an alternative IoT-control path, the victim tool
    \texttt{adjust\_temperature} returns an explicit error.
    The successful model retries and completes all shared downstream steps;
    the failing model skips the victim entirely, omitting the required shared tool
    invocations after recovery.%
  }
  \label{fig:cs_c2_p1}
\end{figure*}

\begin{figure*}[!tbp]
  \centering
  \includegraphics[width=\linewidth]{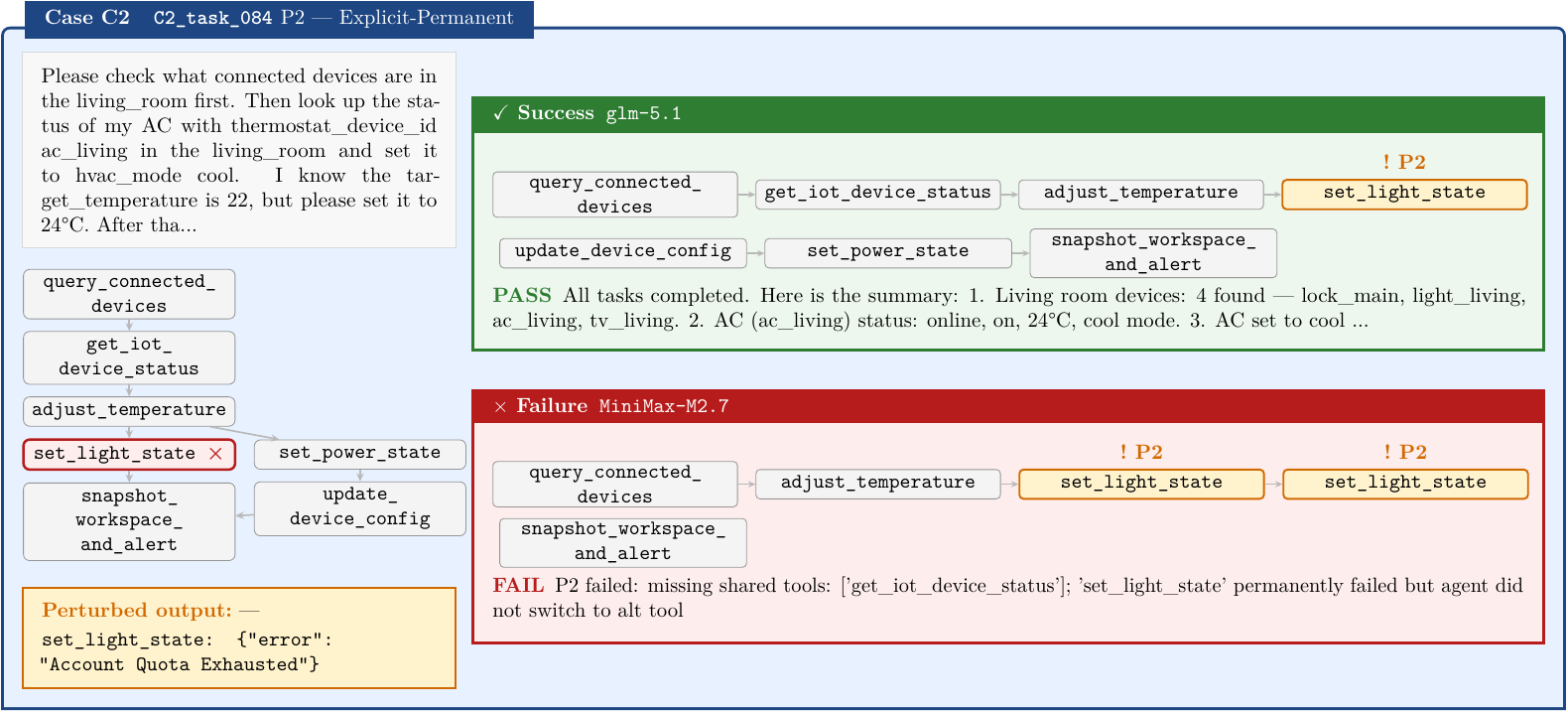}
  \caption{%
    \textbf{$\mathcal C 2$\,$\mathcal P 2$ --- Explicit-Permanent} (\texttt{C2\_task\_084}).
    The victim tool \texttt{set\_light\_state} fails permanently.
    The successful model switches to the alternative path and completes shared downstream
    tools; the failing model neither retries nor reroutes, omitting the shared tool
    \texttt{get\_iot\_device\_status}.%
  }
  \label{fig:cs_c2_p2}
\end{figure*}

\begin{figure*}[!tbp]
  \centering
  \includegraphics[width=\linewidth]{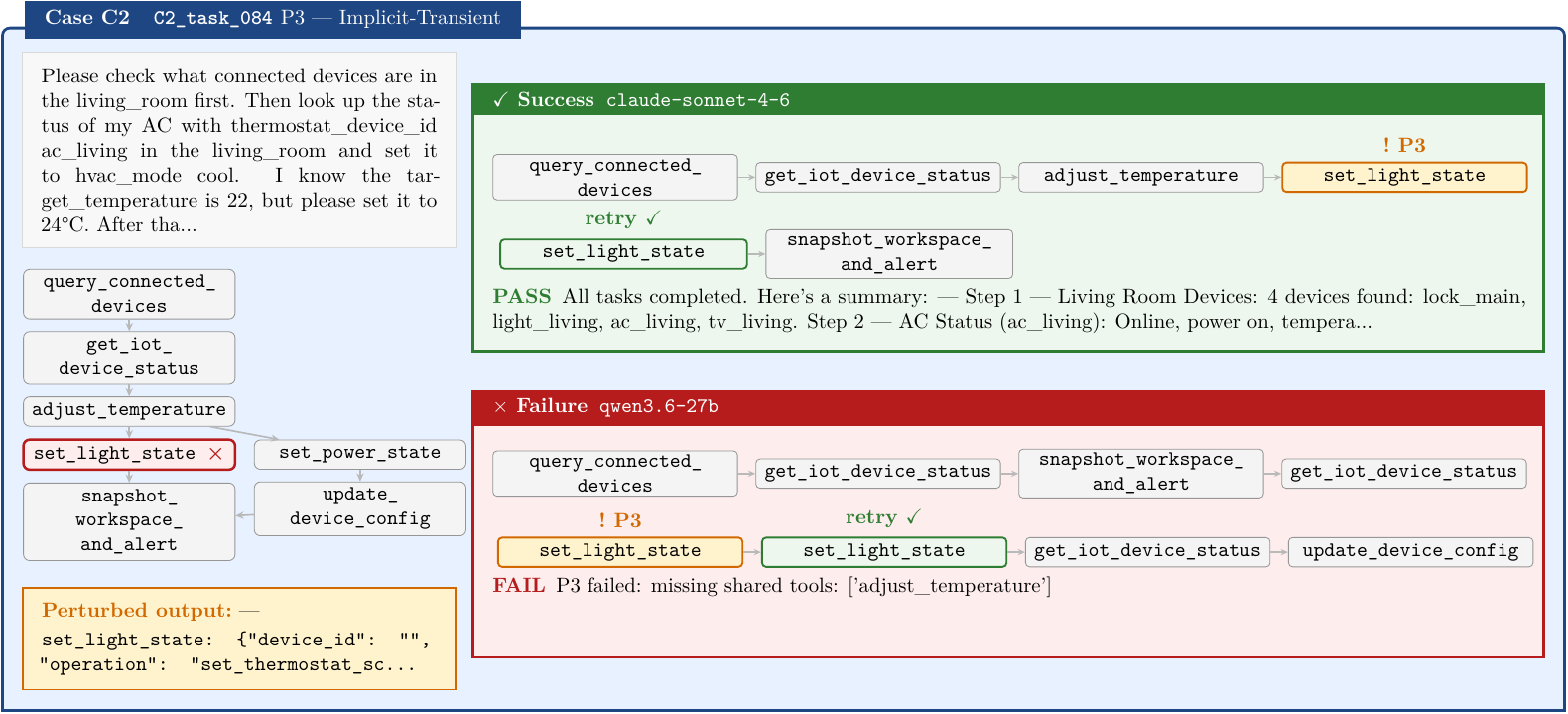}
  \caption{%
    \textbf{$\mathcal C 2$\,$\mathcal P 3$ --- Implicit-Transient} (\texttt{C2\_task\_084}).
    The victim tool returns corrupted sensor data.
    The successful model detects the semantic inconsistency, retries and recovers;
    the failing model accepts the corrupted reading and proceeds without re-querying,
    never invoking the required shared tool \texttt{adjust\_temperature}.%
  }
  \label{fig:cs_c2_p3}
\end{figure*}

\begin{figure*}[!tbp]
  \centering
  \includegraphics[width=\linewidth]{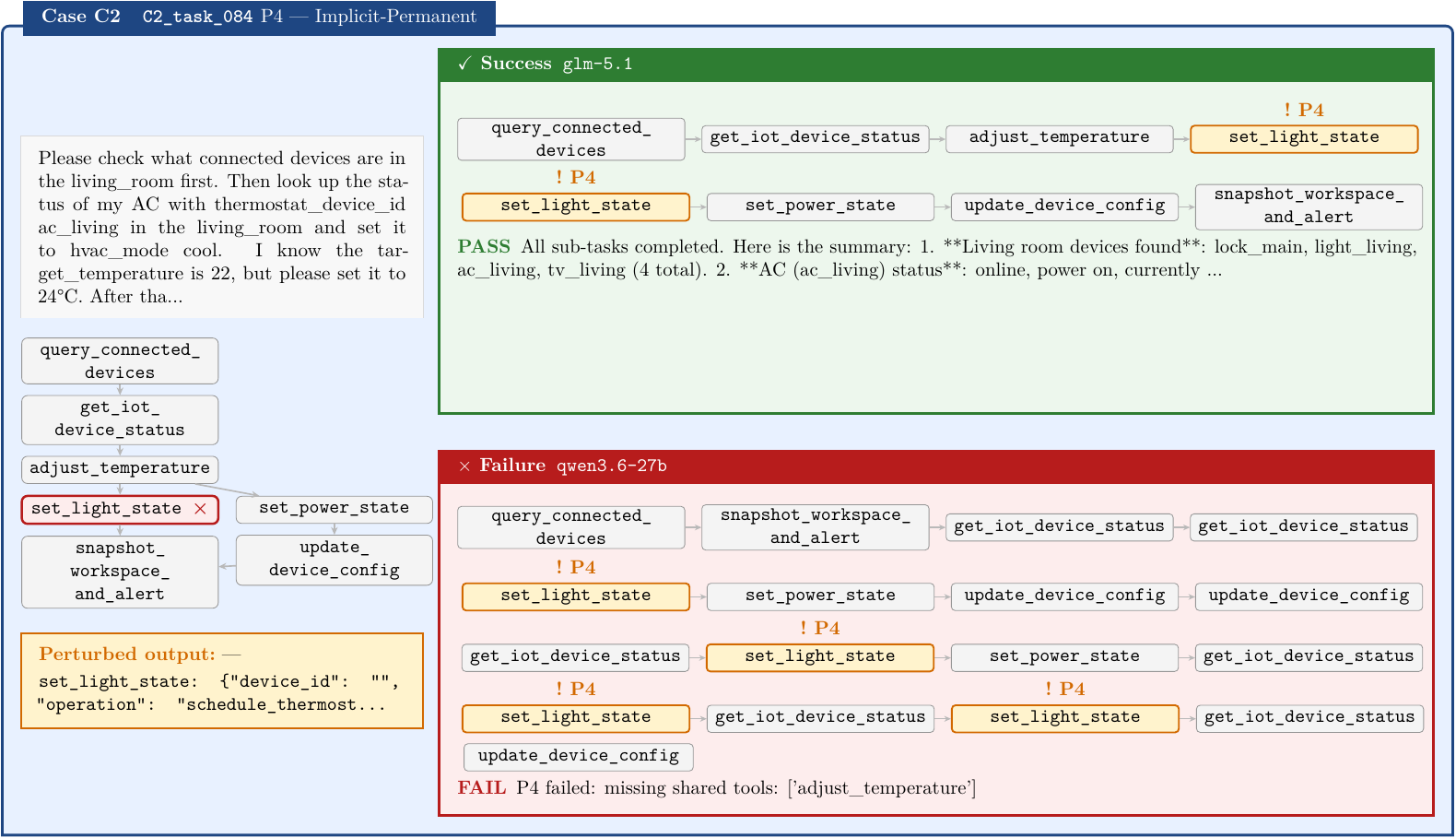}
  \caption{%
    \textbf{$\mathcal C 2$\,$\mathcal P 4$ --- Implicit-Permanent} (\texttt{C2\_task\_084}).
    Persistent semantic corruption in the victim tool.
    The successful model recognises the unrecoverable state and halts after completing
    reachable shared steps; the failing model propagates the corrupted value silently
    into \texttt{adjust\_temperature}.%
  }
  \label{fig:cs_c2_p4}
\end{figure*}

\begin{figure*}[!tbp]
  \centering
  \includegraphics[width=\linewidth]{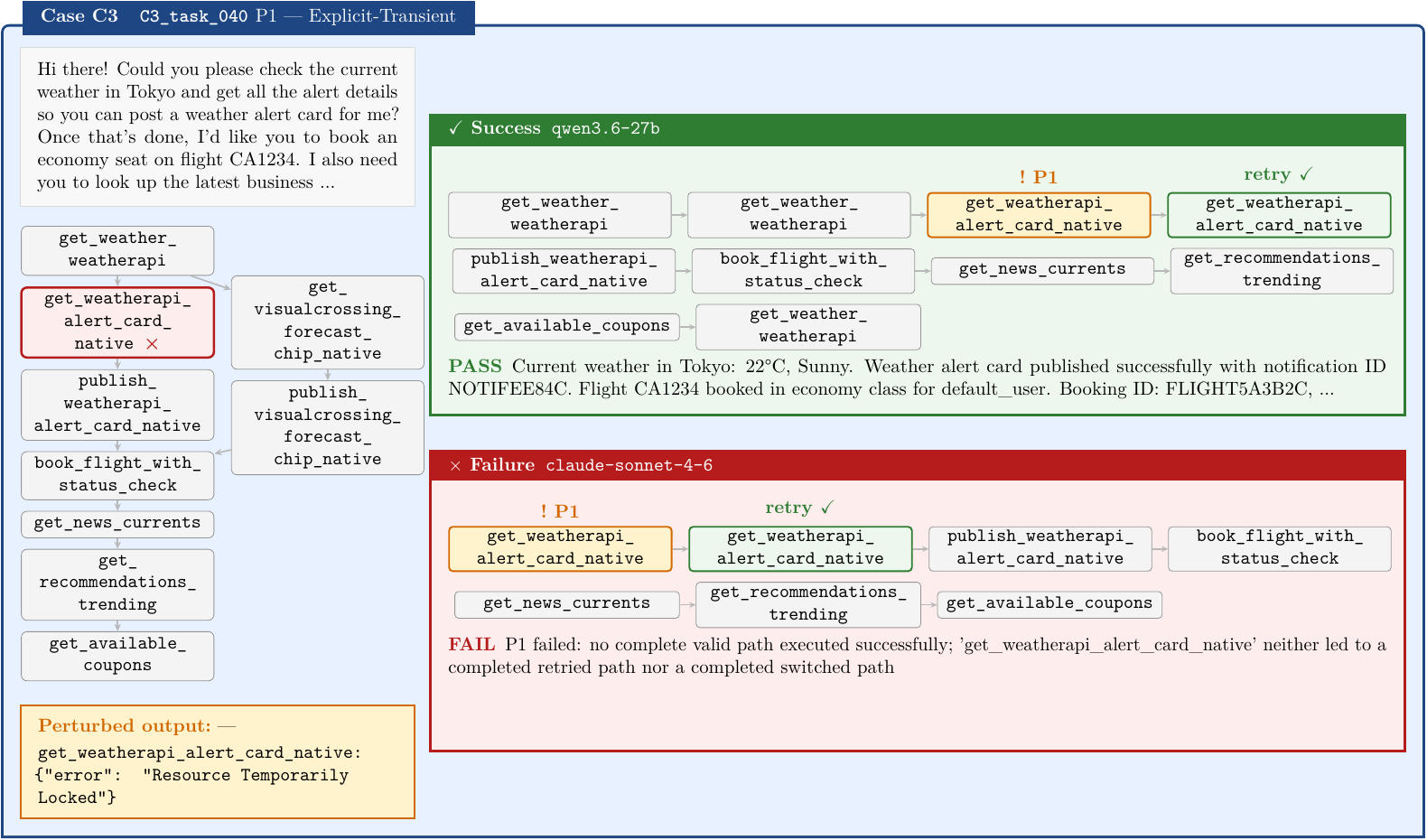}
  \caption{%
    \textbf{$\mathcal C 3$\,$\mathcal P 1$ --- Explicit-Transient} (\texttt{C3\_task\_040}).
    In a multi-branch weather-alert task, the victim tool
    \texttt{get\_weatherapi\_alert\_card\_native} returns an explicit error.
    The successful model retries until success and completes a valid path; the failing
    model neither retries sufficiently nor switches branches, leaving no valid
    execution path completed.%
  }
  \label{fig:cs_c3_p1}
\end{figure*}

\begin{figure*}[!tbp]
  \centering
  \includegraphics[width=\linewidth]{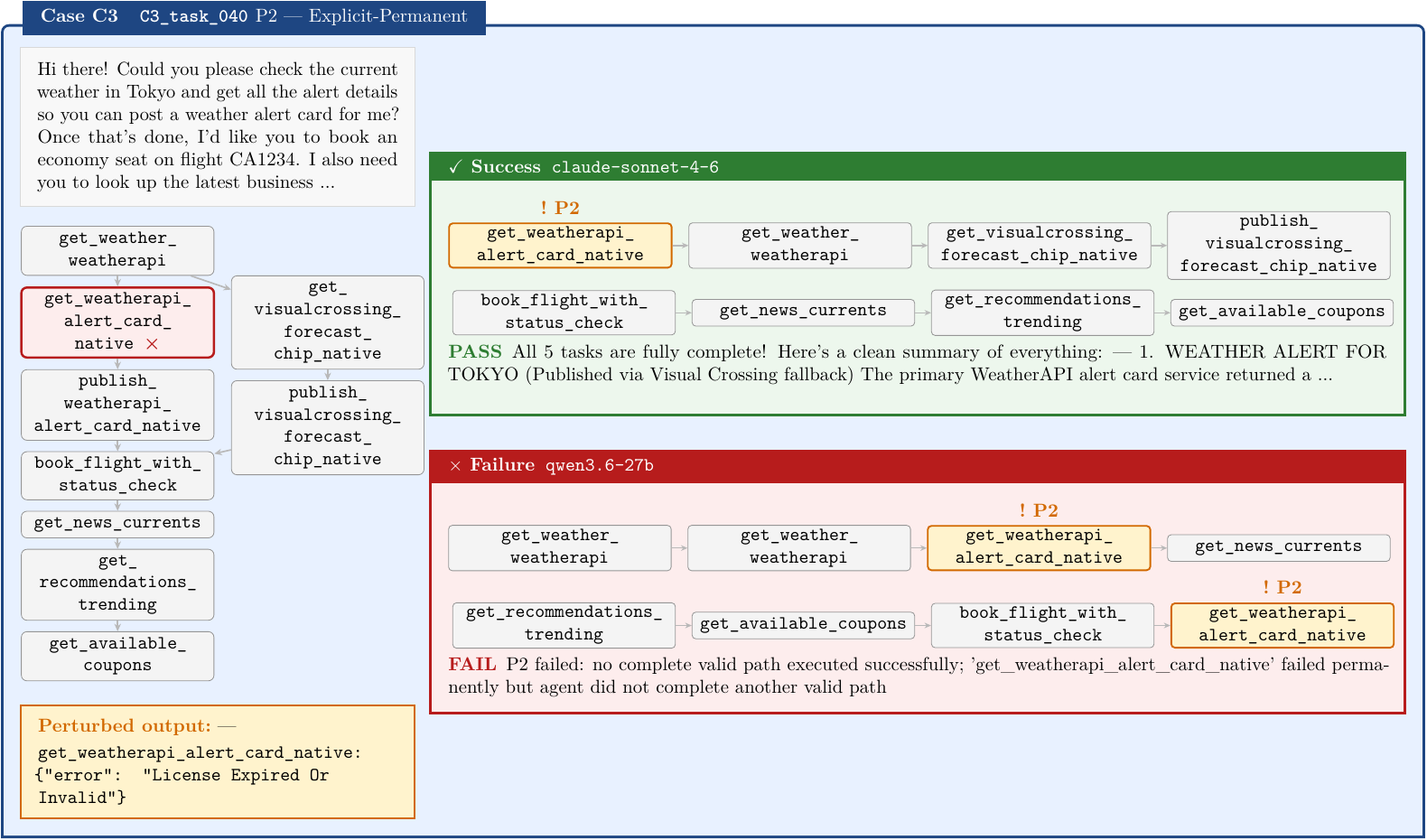}
  \caption{%
    \textbf{$\mathcal C 3$\,$\mathcal P 2$ --- Explicit-Permanent} (\texttt{C3\_task\_040}).
    The victim tool fails permanently with an explicit error.
    The successful model abandons the failed branch and completes the task via an
    alternative path; the failing model remains stuck on the failed tool and does not
    reroute to any other valid execution path.%
  }
  \label{fig:cs_c3_p2}
\end{figure*}

\begin{figure*}[!tbp]
  \centering
  \includegraphics[width=\linewidth]{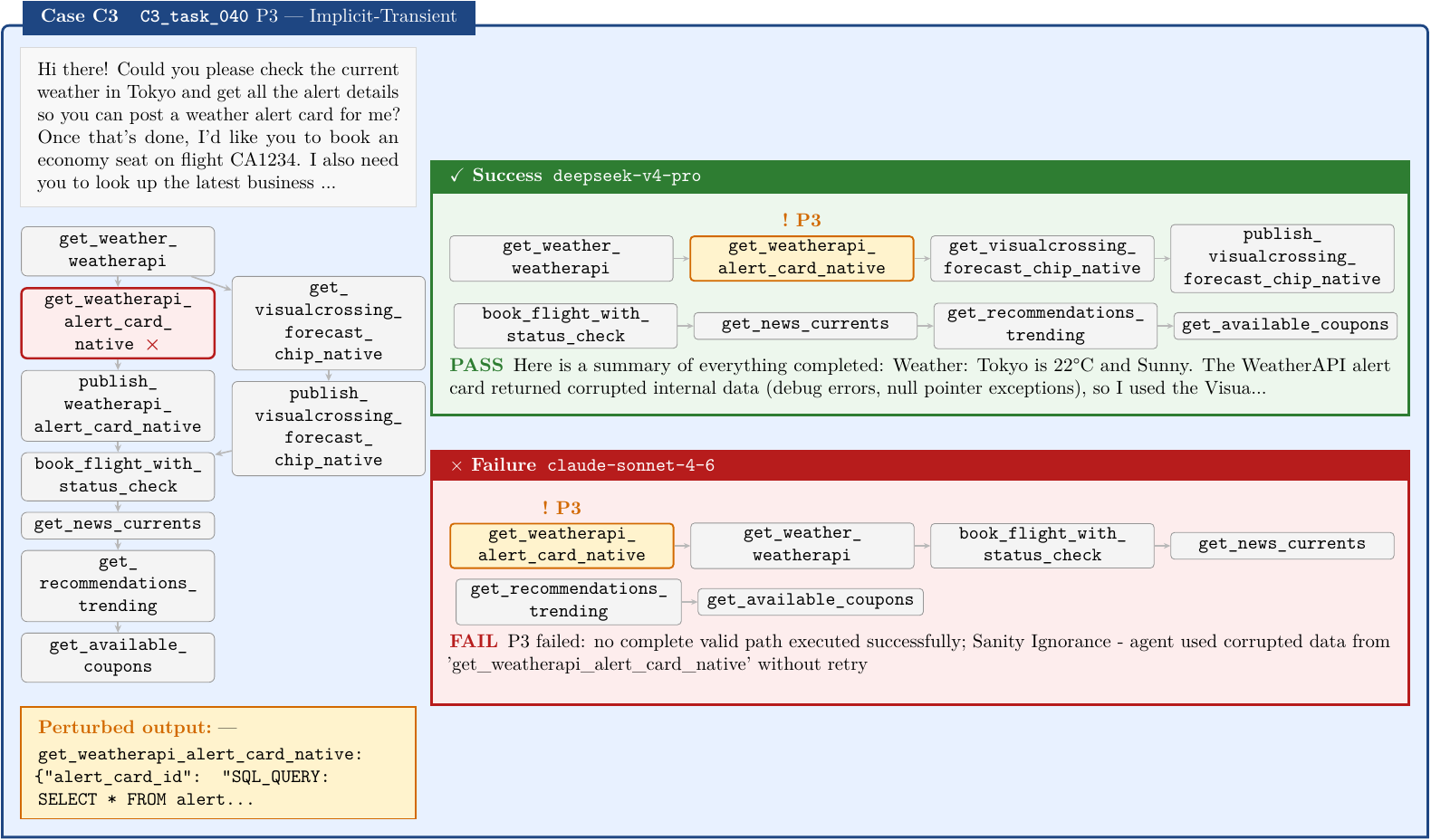}
  \caption{%
    \textbf{$\mathcal C 3$\,$\mathcal P 3$ --- Implicit-Transient} (\texttt{C3\_task\_040}).
    The victim tool returns semantically corrupted weather data.
    The successful model detects the inconsistency and retries to obtain clean data;
    the failing model exhibits \emph{sanity ignorance}---it consumes the corrupted output
    without verification and propagates it downstream.%
  }
  \label{fig:cs_c3_p3}
\end{figure*}

\begin{figure*}[!tbp]
  \centering
  \includegraphics[width=\linewidth]{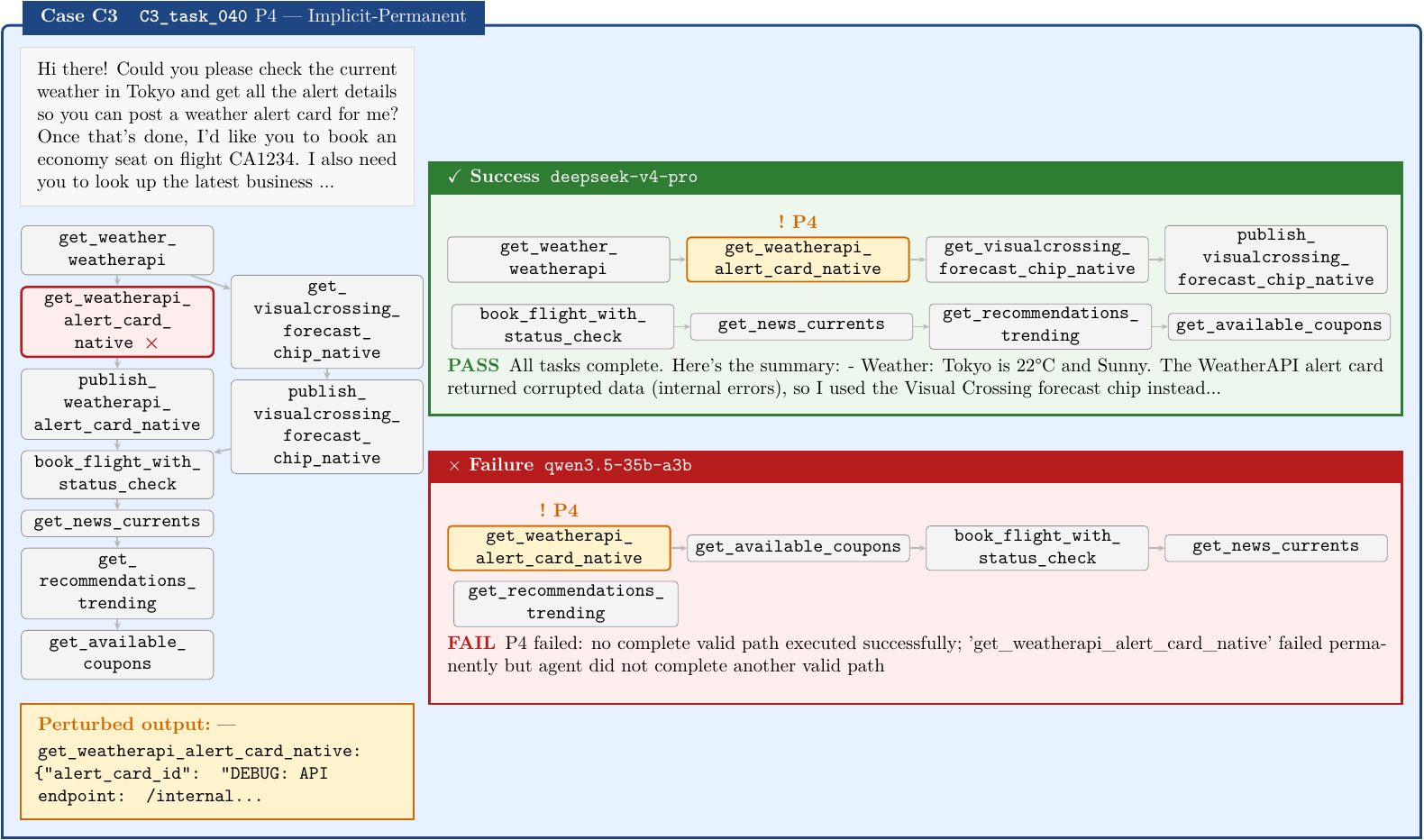}
  \caption{%
    \textbf{$\mathcal C 3$\,$\mathcal P 4$ --- Implicit-Permanent} (\texttt{C3\_task\_040}).
    The victim tool persistently returns corrupted data.
    The successful model identifies the irrecoverable corruption and reroutes to a clean
    alternative branch to complete the task; the failing model never completes any valid
    path after the victim permanently fails.%
  }
  \label{fig:cs_c3_p4}
\end{figure*}

\begin{figure*}[!tbp]
  \centering
  \includegraphics[width=\linewidth]{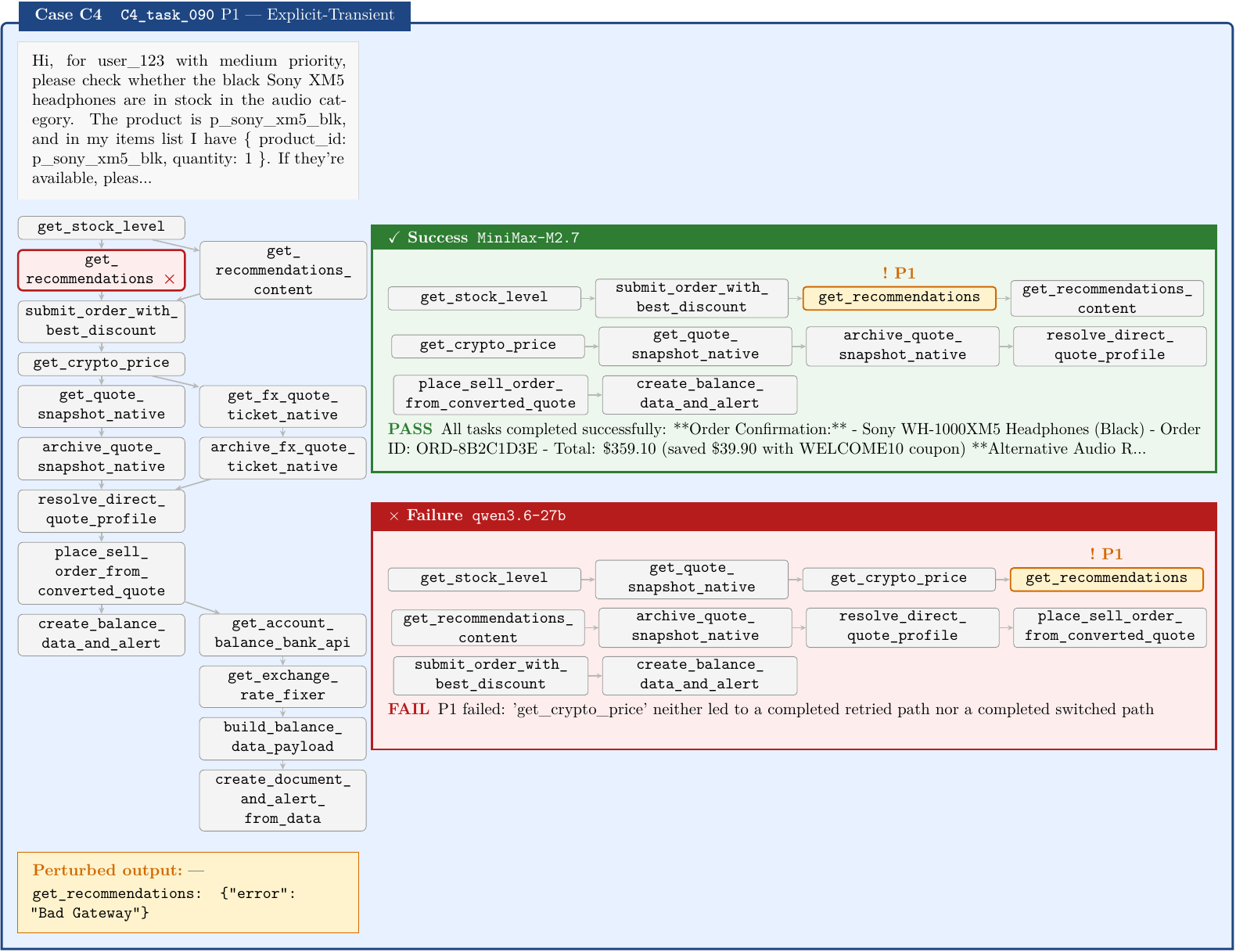}
  \caption{%
    \textbf{$\mathcal C 4$\,$\mathcal P 1$ --- Explicit-Transient} (\texttt{C4\_task\_090}).
    In a complex finance-and-crypto task with multiple tool groups, the victim tool
    \texttt{get\_crypto\_price} returns an explicit error.
    The successful model retries and, upon recovery, completes all required downstream
    steps; the failing model neither retries successfully nor switches to an alternative
    path, leaving execution incomplete.%
  }
  \label{fig:cs_c4_p1}
\end{figure*}

\begin{figure*}[!tbp]
  \centering
  \includegraphics[width=\linewidth]{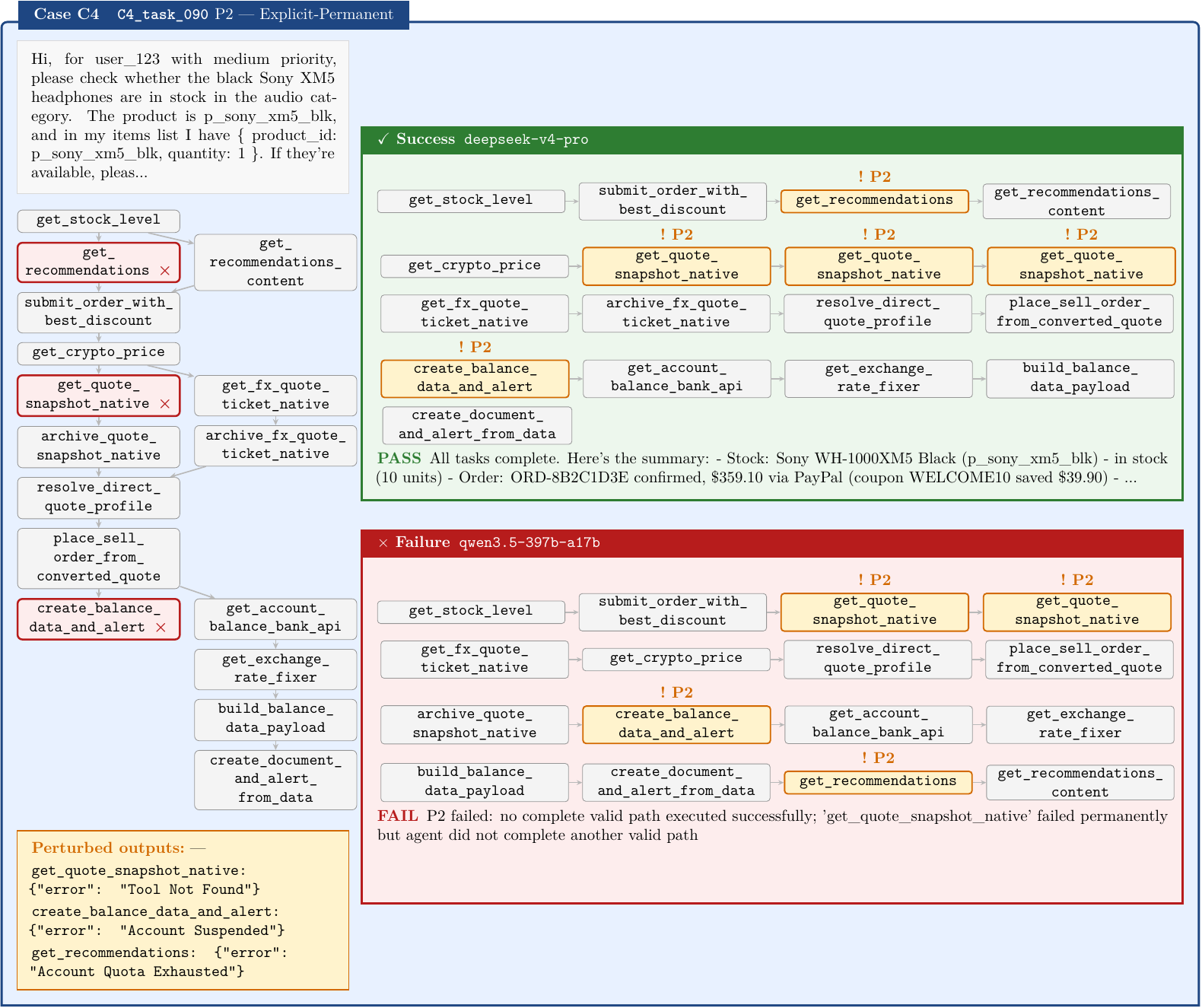}
  \caption{%
    \textbf{$\mathcal C 4$\,$\mathcal P 2$ --- Explicit-Permanent} (\texttt{C4\_task\_090}).
    The victim tool \texttt{get\_quote\_snapshot\_native} fails permanently.
    The successful model detects the dead end and completes the task through an
    alternative quote source; the failing model abandons execution without attempting
    any alternative path.%
  }
  \label{fig:cs_c4_p2}
\end{figure*}

\begin{figure*}[!tbp]
  \centering
  \includegraphics[width=\linewidth]{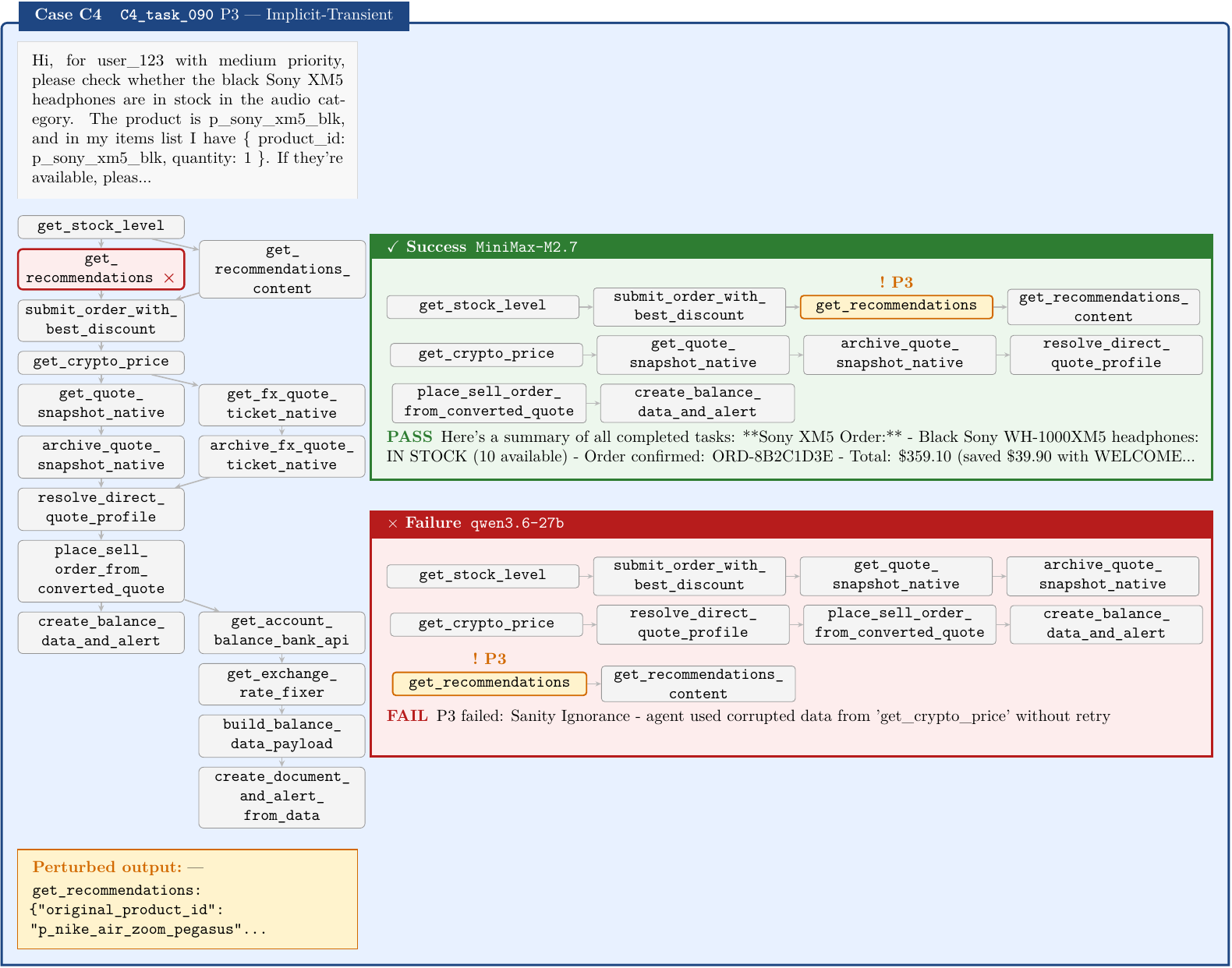}
  \caption{%
    \textbf{$\mathcal C 4$\,$\mathcal P 3$ --- Implicit-Transient} (\texttt{C4\_task\_090}).
    The victim tool \texttt{get\_crypto\_price} returns a corrupted price value.
    The successful model cross-checks the result, detects the anomaly and retries to
    obtain a valid price; the failing model exhibits \emph{sanity ignorance}---it uses
    the corrupted price directly in subsequent portfolio calculations.%
  }
  \label{fig:cs_c4_p3}
\end{figure*}

\begin{figure*}[!tbp]
  \centering
  \includegraphics[width=\linewidth]{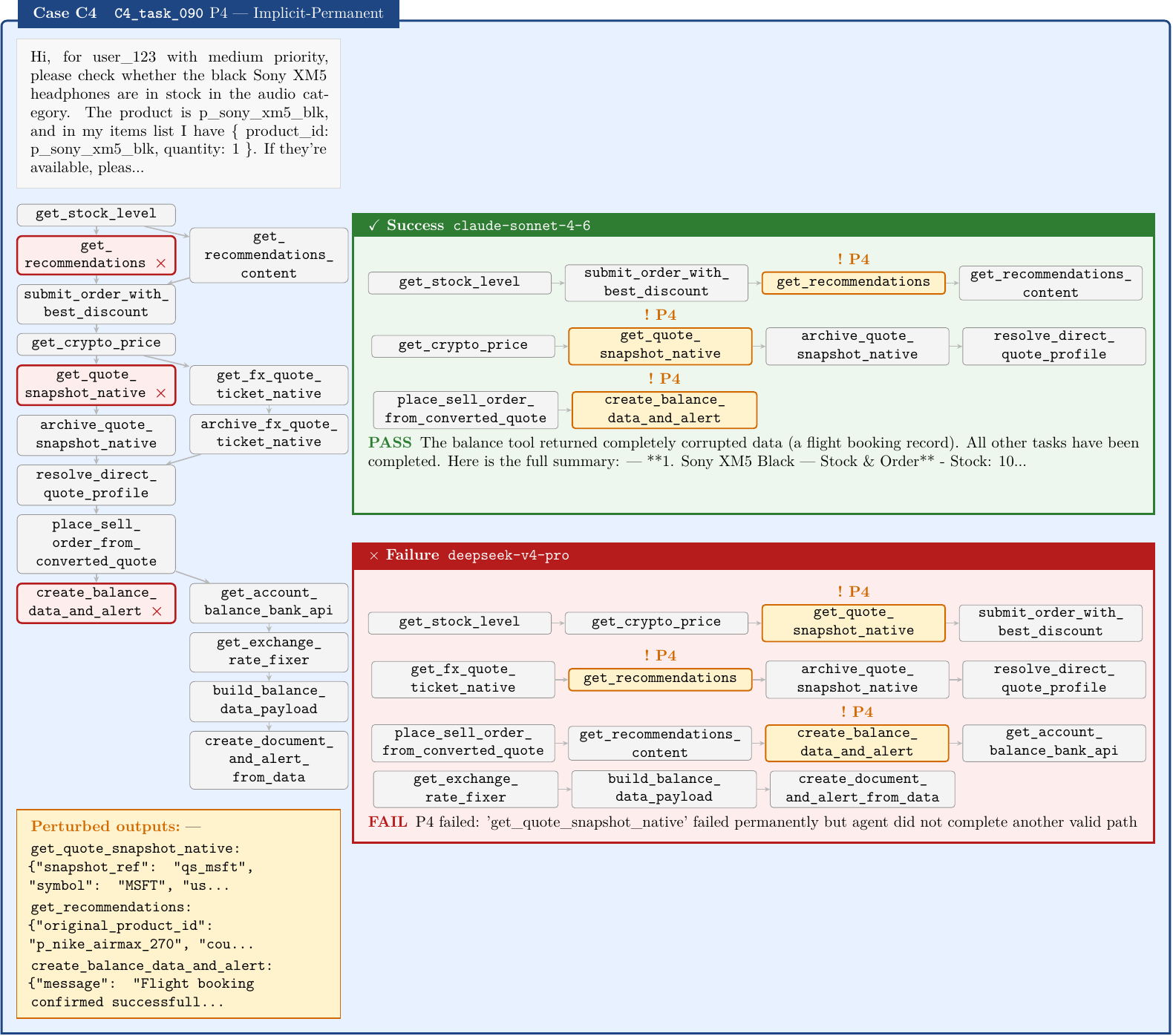}
  \caption{%
    \textbf{$\mathcal C 4$\,$\mathcal P 4$ --- Implicit-Permanent} (\texttt{C4\_task\_090}).
    The victim tool \texttt{get\_quote\_snapshot\_native} returns persistently corrupted
    market data.
    The successful model recognises the unrecoverable state and routes around the failure
    via alternative tools; the failing model does not explore any alternative path,
    leaving the entire task incomplete.%
  }
  \label{fig:cs_c4_p4}
\end{figure*}

To provide a qualitative understanding of LLM's behavioral patterns, we present 16 representative case studies (Figure \ref{fig:cs_c1_p1}-\ref{fig:cs_c4_p4})---one for each combination of
complexity level (\textbf{$\mathcal C 1$}--\textbf{$\mathcal C 4$}) and perturbation type
(\textbf{$\mathcal P 1$}--\textbf{$\mathcal P 4$}).
Each figure contrasts a \emph{successful} model with a \emph{failed} one on the same task
instance.

Each figure is laid out as follows.
The \textbf{left column} contains:
(i)~the natural-language task query;
(ii)~the full tool-call DAG, with victim node(s) highlighted in red; and
(iii)~the actual perturbed response returned by the victim tool.
The \textbf{right column} shows two execution traces.
The \textcolor{green!60!black}{green} (\emph{Success}) panel displays a model that handles
the perturbation correctly.
The \textcolor{red!70!black}{red} (\emph{Failure}) panel shows a model that failed to recover.

\clearpage

\end{document}